\documentclass[runningheads]{llncs}

\usepackage[mobile]{eccv}

\usepackage{eccvabbrv}

\usepackage{graphicx}
\usepackage{booktabs}
\usepackage{multirow}
\usepackage{amssymb}
\usepackage{pifont}
\usepackage{bbm}

\usepackage{graphicx}
\newsavebox\CBox
\def\textBF#1{\sbox\CBox{#1}\resizebox{\wd\CBox}{\ht\CBox}{\textbf{#1}}}

\usepackage{amsmath}

\DeclareMathOperator*{\argmin}{arg\,min}

\usepackage{numprint}

\def\littlesection#1{\smallskip\noindent\textbf{#1}}

\usepackage[accsupp]{axessibility}  % Improves PDF readability for those with disabilities.

\usepackage{hyperref}

\usepackage{orcidlink}

\begin{document}

\title{OGNI-DC: Robust Depth Completion with Optimization-Guided Neural Iterations}

\titlerunning{OGNI-DC: Depth Completion with Optimization-Guided Iterations}

\author{Yiming Zuo \and
Jia Deng}

\authorrunning{Y.~Zuo and J.~Deng}

\institute{Department of Computer Science, Princeton University \\
\email{\{zuoym,jiadeng\}@princeton.edu}}

\maketitle

\begin{abstract}
  Depth completion is the task of generating a dense depth map given an image and a sparse depth map as inputs. It has important applications in various downstream tasks. In this paper, we present OGNI-DC, a novel framework for depth completion. The key to our method is ``\textbf{O}ptimization-\textbf{G}uided \textbf{N}eural \textbf{I}terations'' (OGNI). It consists of a recurrent unit that refines a depth gradient field and a differentiable depth integrator that integrates the depth gradients into a depth map. OGNI-DC exhibits strong generalization, outperforming baselines by a large margin on unseen datasets and across various sparsity levels. Moreover, OGNI-DC has high accuracy, achieving state-of-the-art performance on the NYUv2 and the KITTI benchmarks. Code is available at \url{https://github.com/princeton-vl/OGNI-DC}.
  
\end{abstract}

\section{Introduction}
\label{sec:intro}

Depth completion is the task of predicting a pixel-wise depth map from a single RGB image and known sparse depth. The sparse depth can come from depth sensors such as Lidar~\cite{kittidc,ddad} and structured light~\cite{nyuv2}, or multiview systems such as SLAM or Structure-from-Motion~\cite{void}. Depth completion has important applications such as autonomous driving~\cite{carranza2022object,hane20173d}, robotics~\cite{liao2017parse}, and augmented reality~\cite{holynski2018fast}. \looseness=-1

For a depth completion system to be maximally useful, it needs to be not only accurate but also robust, meaning that it continues to perform well even with large distribution shifts in input, such as the type of scenes and the sparsity patterns of known depth. Such robustness is desirable because it allows a single system to perform well across a wide range of conditions. 

However, achieving both accuracy and robustness at the same time remains challenging. Early works~\cite{hawe2011dense,liu2015depth,zhang2018deep} on depth completion formulate it as an optimization problem with hand-crafted energy terms. Specifically, Zhang \etal~\cite{zhang2018deep} propose to solve a global optimization problem, where the depth needs to agree with the sparse observations at valid locations while being constrained by surface normals and local smoothness. While optimization-based approaches exhibit strong cross-dataset generalization, they are not accurate enough. 

More recent methods are based on deep learning~\cite{cheng2020cspn++,park2020non,xu2020deformable,lin2022dynamic,zhang2023completionformer,zhou2023bev} and usually directly regress depth values with deep neural networks. Such methods achieve impressive accuracy on benchmarks such as NYUv2~\cite{nyuv2} and KITTI~\cite{kittidc} when trained with data of the same domain. However, the models are not robust to shifts in scene distributions or sparsity patterns of known depth, and often fail catastrophically when tested on new datasets, especially those with different depth ranges or sparsity levels~\cite{conti2023sparsity,bartolomei2023revisiting}.

In this paper, we propose a novel depth completion method, OGNI-DC, which achieves both the superior accuracy of deep neural networks and the robustness of optimization-based approaches. The core of our method is ``\textbf{O}ptimization-\textbf{G}uided \textbf{N}eural \textbf{I}terations'' (OGNI); it consists of a recurrent network that iteratively refines a field of \textit{depth gradients} (\ie, depth differences between neighboring pixels) and a novel \textit{Differentiable Depth Integrator} (DDI) that integrates the depth gradients into a dense depth map. In OGNI-DC, refinement and integration are tightly coupled, meaning that the neural refinement of the depth gradients depends on the current depth integration result, and depth integration is guided by the neural refinement.

\begin{figure}[t]
  \centering
  \includegraphics[width=\linewidth]{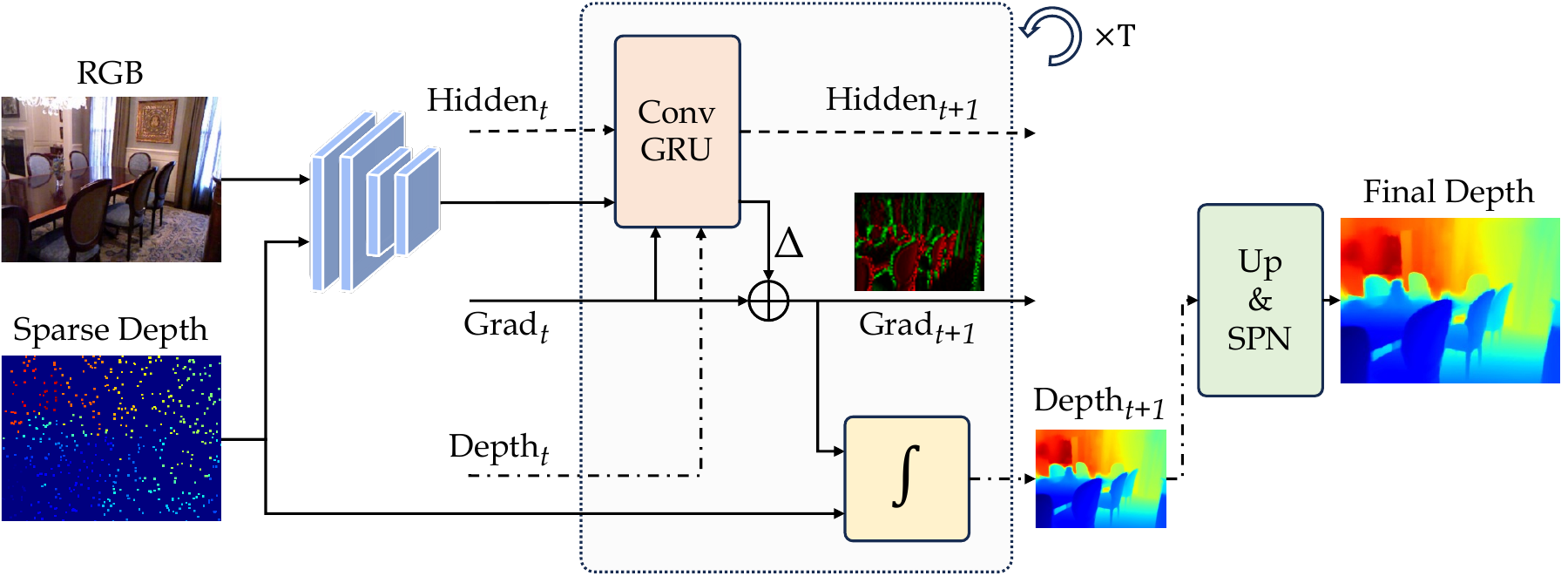}
  \caption{The overall pipeline of OGNI-DC. We first extract features at $1/4$ resolution from the concatenation of the image and the sparse depth map. After that, a ConvGRU iteratively refines a depth gradient field based on current predictions, and the DDI (\cref{sec:DDI}) integrates the depth gradient field into an intermediate depth map. Finally, we up-sample the intermediate depth map and enhance it with a Spatial Propagation Network (SPN)~\cite{lin2022dynamic} to get the full-resolution depth map.
  }
  \label{fig:pipeline}
\end{figure}

DDI is the key to our design. DDI integrates the depth gradients into a depth map while satisfying the boundary conditions set by the sparse depth observations. DDI is \textit{differentiable}, meaning that the errors on the predicted depth map can be back-propagated to the depth gradients, making the full pipeline end-to-end trainable. 

DDI improves robustness by exploiting the fact that depth can be recovered from depth gradients and known sparse depth through optimization. Such optimization explicitly constrains the depth to be consistent with the sparse depth, and thus enables the model to easily adapt to varying patterns of sparse depth observations without retraining. In addition, unlike metric depth prediction which usually requires reasoning on the entire image, depth gradients can often be inferred from a local window and are thus easier to predict. An easier learning task leads to better generalization given the same amount of training data. \looseness=-1

Another important design is recurrent refinement through a convolutional gated recurrent unit (ConvGRU). We feed the integration result from the previous step back to the ConvGRU, which then refines the depth gradients. The recurrent refinement makes the network aware of the consequences of its depth gradients outputs and thus provides stronger guidance and regularization.

The design of OGNI-DC is substantially different from the previous deep-learning-based depth completion methods. First, instead of directly predicting depth, OGNI-DC predicts depth gradients, which is equally expressive but easier to learn. Second, the constraints on depth from the sparse depth observations are explicitly enforced, rather than from a brittle, learned identity mapping~\cite{park2020non,zhang2023completionformer}. 

The high-level idea of OGNI-DC draws inspiration from previous works that use coupled optimization and iterative refinement, such as DROID-SLAM~\cite{teed2021droid} and DPVO~\cite{teed2024deep}. However, prior works were limited to multiview tasks, and we are the first to apply it to a single-view task. Therefore, our designs of the optimization layer are completely different: our optimization enforces constraints on depth and depth gradients from a single image, whereas prior works enforce constraints on depth and pixel correspondences across multiple views.  

To prove the effectiveness of our system, we conduct extensive experiments on common benchmarks. We evaluate the model's accuracy under both the zero-shot generalization setting on the DADD~\cite{ddad} and the VOID~\cite{void} datasets, as well as the in-domain setting on the NYUv2~\cite{nyuv2} and the KITTI~\cite{kittidc} datasets.

OGNI-DC exhibits strong generalization. When trained on NYUv2~\cite{nyuv2} and tested on VOID\cite{void}, our model reduces MAE by 35.5\% compared to prior works.
Similarly, when trained on KITTI~\cite{kittidc} alone, our model reduces RMSE by 25.0\% on DDAD~\cite{ddad} from the best previous model trained on the same data. OGNI-DC is also robust to the density changes of the sparse depth observations. On NYUv2~\cite{nyuv2}, our one single model works the best across a wide density range of $50\sim\numprint{20000}$ points. On KITTI~\cite{kittidc}, our model trained with 64 Lidar scanning lines outperforms previous methods by a large margin when tested with $8\sim32$ lines, reducing MAE by 25.5\% on the 16-Lines input.

Finally, OGNI-DC achieves state-of-the-art in-domain accuracy. It achieves the best performance on the NYUv2~\cite{nyuv2} benchmark, improving RMSE from 0.089m to 0.087m. On the KITTI~\cite{kittidc} online benchmark, OGNI-DC outperforms all previous methods on the MAE, iRMSE, and iMAE metrics when trained with an ${L_1}$ loss.

\section{Related Works}
\label{sec:related_works}
\subsection{Depth Completion}

Deep learning-based methods have achieved impressive accuracy for depth completion. Early works directly predict depth values from image and sparse depth inputs~\cite{kittidc,liu2021fcfr,ma2018sparse,tang2019learning}. These methods are feed-forward and have difficulties in reasoning long-range depth dependencies because of the limited receptive field.

To mitigate this issue, more recent works~\cite{cheng2019cspn,cheng2020cspn++,cheng2019learning,hu2021penet,liu2022graphcspn,park2020non,xu2020deformable,lin2022dynamic} propose a series of Spatial Propagation Networks (SPNs). SPNs iteratively update the regressed depth map based on the predictions of each pixel and its neighbors. Specifically, NLSPN~\cite{park2020non} predicts a non-local neighborhood by utilizing deformable convolutions. DySPN~\cite{lin2022dynamic} predicts different propagation weights for each iteration and achieves better performance with fewer iterations and neighbors. BEV@DC~\cite{zhou2023bev} uses 3D SPN on the unprotected Lidar point cloud at training time to regularize the solution space. Although SPN-based method and OGNI-DC both involve iterative updates, they are significantly different: the SPN-based method predicts depth values and propagation guidance only once, and uses SPN to propagate depth predictions without explicit constraints around observed locations. In contrast, our method predicts depth gradients iteratively based on current depth, and explicitly enforces the observation constraints through optimization. 

A recent work LRRU~\cite{wang2023lrru} generates an initial dense depth map with handcrafted heuristics instead of neural networks, and propagates the depth values iteratively through spatially-variant kernels. While both LRRU and our method avoid direct depth regression, we are different in: 1) LRRU is an SPN variant that directly propagates depth values, while our recurrent unit performs updates on the depth gradients and is coupled with DDI to produce depth outputs. 2) The updates in LRRU are coarse-to-fine instead of recurrent, only being applied once at each resolution, whereas our recurrent update can be unrolled arbitrary times at the same resolution. 

Several works focus on generalization. SpAgNet~\cite{conti2023sparsity} merges the sparse observations into the multi-resolution depth maps predicted by a network and can deal with extremely sparse inputs. Concurrent work SparseDC~\cite{long2023sparsedc} designs an enhanced backbone and an uncertainty-based fusion module to make the network robust to depth density changes. VPP4DC~\cite{bartolomei2023revisiting} repurposes a stereo matching network for depth completion by projecting random patterns onto a virtual neighboring view, and achieves decent zero-shot performance on the DDAD~\cite{ddad} and the VOID~\cite{void} datasets. The generalization of all these methods comes with a cost of accuracy, as none of these works achieves comparable performance on NYUv2~\cite{nyuv2} or KITTI~\cite{kittidc} as ours. Furthermore, some methods focus on cross-dataset generalization, while others focus on cross-density generalization, but none of these methods achieve satisfactory results in both scenarios.

\subsection{Geometry Reconstruction from Local Properties}

Some prior works propose to solve 3D vision tasks with constraints from local properties, such as surface normals~\cite{zhao2021confidence,hu2023surface,qi2018geonet,yin2021virtual,long2021adaptive,qiu2019deeplidar}, occlusion boundaries~\cite{zhang2018deep,ramamonjisoa2020predicting,imran2021depth}, and principle directions~\cite{huang2019framenet}. Such local properties are easier to learn and generalize better to unseen domains, compared to global properties such as depth~\cite{zhang2018deep}. Here we mainly introduce prior works on monocular depth estimation, as it is the most closely related task. 
Long \etal~\cite{long2021adaptive} employs the depth-normal constraints by sampling reliable planar regions in the pixel space. DeepLiDAR~\cite{qiu2019deeplidar} predicts surface normals as an intermediate representation and converts it to depth with a neural network. Compared to them, OGNI-DC predicts depth gradients instead of normals, as depth gradients are defined everywhere and are capable of reconstructing scenes with many thin structures such as trees and fences, whereas surface normals are not defined on occlusion boundaries. While depth gradients have been used in previous works as a loss term~\cite{li2018megadepth,Ranftl2022midas}, we are the first to directly predict depth gradients from the network.\looseness=-1

Some previous methods solve geometry reconstruction problems with iterative updates~\cite{teed2020raft,lipson2021raft,qi2020geonet++,bae2022irondepth,shao2023nddepth,teed2021droid}. Among them,
GeoNet++~\cite{qi2020geonet++} iteratively refines normal-from-depth and depth-from-normal. IronDepth~\cite{bae2022irondepth} proposes depth propagation candidates based on local planes defined by the normals.
NDDepth~\cite{shao2023nddepth} performs contrastive iterative updates on the two depth maps directly predicted and computed from normals. Compared to ours, none of the iterative units of these methods involve an optimization-based layer, and cannot be easily extended to the depth completion task.

Several works utilize optimization-inspired designs to solve various computer vision tasks~\cite{yeh2022total, bai2022deep,hu2023surface,liu2023va}. Specifically, VA-DepthNet~\cite{liu2023va} solves an integration problem in the feature space at 1/16 resolution. Compared to them, our DDI layer solves an optimization problem at a larger scale directly in the depth space.

Some other methods employ coupled iterative refinement and differentiable optimization~\cite{teed2021droid,teed2024deep}. DROID-SLAM~\cite{teed2021droid} iteratively updates the optical flow, and uses a Dense Bundle Adjustment (DBA) layer to optimize camera poses and depths. Compared to DROID-SLAM, the nature of the tasks and the designs of our differentiable optimization layers are different. The DBA layer solves a non-linear optimization problem and performs only two Gauss-Newton steps at each iteration. Our DDI layer solves a linear least-squares problem and finds the global minimum through the conjugate gradient method~\cite{hestenes1952methods}. This difference is non-trivial and poses additional challenges in the backward pass. \looseness=-1

\vspace{-2mm}
\section{Approach Overview}
\label{sec:pipeline}

In this section, we describe our depth completion pipeline. Our model takes an RGB image $\mathbf{I} \in \mathbb{R}^{3 \times H \times W}$, a sparse depth observation map $\mathbf{O} \in \mathbb{R}_+^{H \times W}$, and a mask $\mathbf{M} \in \{0,1\}^{H \times W}$ as input, where $\mathbf{M}_{i,j} = 1$ if and only if there is a valid observation at location ($i,j$). Our model outputs a dense depth map $\mathbf{\hat{D}} \in \mathbb{R}_+^{H \times W}$. 

Our pipeline is illustrated in \cref{fig:pipeline}. There are three main components, \ie, a backbone for feature extraction, a coupled ConvGRU and DDI for intermediate depth maps prediction, and an up-sample and enhancement layer that produces the final depth map. We describe each component below in this section, and leave the details of DDI to \cref{sec:DDI}.

\vspace{-2mm}
\subsection{Feature Extraction}
\label{subsec:backbone}

We use a deep neural network as the backbone to extract features. The input to the backbone is the \textit{concatenation} of the RGB image $\mathbf{I}$ and the sparse observations $\mathbf{O}$. The backbone outputs features at two resolutions: $1/4$ resolution for intermediate depth predictions and full resolution for final depth enhancement:
\begin{equation}
    \mathbf{\hat{F}^{\text{full}}}, \mathbf{\hat{F}^{\frac{1}{4}}} = \operatorname{Backbone}(\mathbf{I}, \mathbf{O}).
\end{equation}

We use CompletionFormer~\cite{zhang2023completionformer} as our backbone for its state-of-the-art performance. We take the decoder features at the full and $1/4$ resolution as outputs.

\subsection{Intermediate Depth Prediction}

We predict $T$ intermediate depth maps in an iterative manner, where $T$ is the total number of steps. In the $t$-th step, we first predict the depth gradients $\mathbf{\hat{G}}_t \in \mathbb{R}^{2 \times \frac{H}{4} \times \frac{W}{4}}$, and then convert $\mathbf{\hat{G}}_t$ into an intermediate depth $\mathbf{\hat{D}^{\frac{1}{4}}}_t \in \mathbb{R}_+^{\frac{H}{4} \times \frac{W}{4}}$ with DDI (see \cref{sec:DDI}). We describe the details of the two modules below.

\littlesection{Depth Gradients Prediction.} This module performs iterative updates on the depth gradients map through a convolutional gated recurrent unit (ConvGRU)~\cite{teed2020raft}. For each step, it predicts a refinement $\Delta \mathbf{\hat{G}}$ which is applied to the current depth gradients: $\mathbf{\hat{G}}_t = 
\Delta \mathbf{\hat{G}} + \mathbf{\hat{G}}_{t-1} $.

This module takes as input the backbone features, the hidden state, the predicted depth gradients, and the predicted intermediate depth map from the previous step. It outputs a depth gradients refinement map and an updated hidden state:
\begin{equation}
    \Delta \mathbf{\hat{G}}, \mathbf{h}_t = \operatorname{ConvGRU}(\mathbf{\hat{F}^{\frac{1}{4}}}, \mathbf{h}_{t-1}, \mathbf{\hat{D}^{\frac{1}{4}}}_{t-1}, \mathbf{\hat{G}}_{t-1}).
\end{equation}

We initialize $\mathbf{\hat{G}}_0$ as an all $\mathbf{0}$ tensor, and $\mathbf{h}_0$ from a 2-layers CNN. $\mathbf{\hat{D}^{\frac{1}{4}}}_{t-1}$ and $\mathbf{\hat{G}}_{t-1}$ are encoded by a 2-layers CNN before feeding into the GRU. The network architecture of ConvGRU is adopted from RAFT~\cite{teed2020raft}.

\littlesection{Gradients Integration.} We use DDI (see \cref{sec:DDI}) to integrate the depth gradients and the down-sampled observations into an intermediate depth map:
\begin{equation}
    \mathbf{\hat{D}^{\frac{1}{4}}}_t = \operatorname{DDI}(\mathbf{\hat{G}}_{t}, \mathbf{O^{\frac{1}{4}}}, \mathbf{M^{\frac{1}{4}}}),
\end{equation}
where $\mathbf{O^{\frac{1}{4}}}$ and $\mathbf{M^{\frac{1}{4}}}$ are obtained by performing an average-pooling on valid pixels from the observation $\mathbf{O}$ and the mask $\mathbf{M}$ respectively.

\begin{figure}[t]
  \centering
  \includegraphics[width=\linewidth]{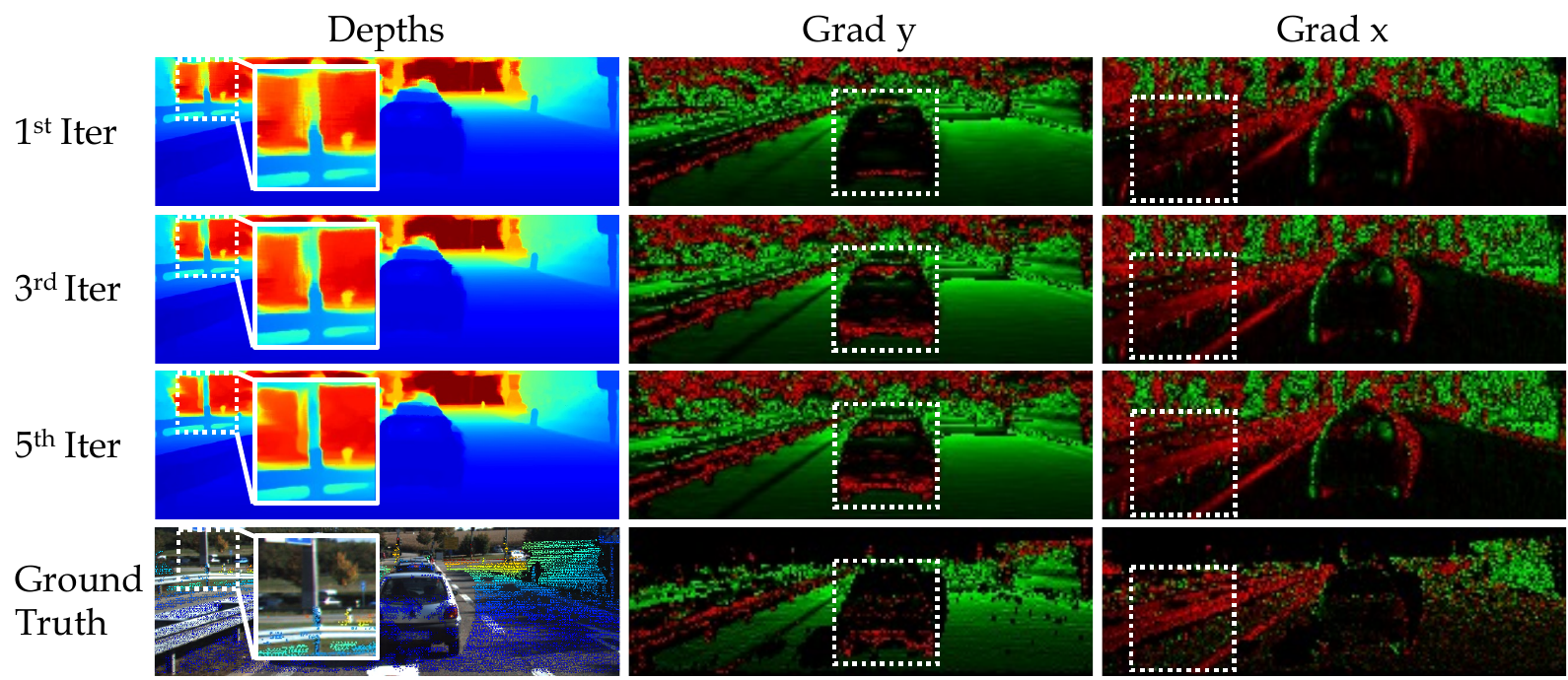}
  \caption{We demonstrate the effectiveness of our iterative refinement. The depth and the gradients predictions in the highlighted areas gradually improved. Red means negative gradients and green means positive gradients. Brighter colors mean larger absolute values. Some ground truths are missing in the gradients map due to incomplete depth.
  }
  \label{fig:iter_refine}
\vspace{-5mm}
\end{figure}

All computations are done at $1/4$ resolution. The intermediate depth prediction involves iterative refinements and differentiable integrations, which put a heavy burden on computation and memory when running at full resolution. Therefore, $1/4$ resolution is a good balance between performance and computational cost. \cref{fig:iter_refine} illustrates the iterative refinement process.

\subsection{Depth Up-sample and Enhancement} We up-sample the intermediate depth map $\mathbf{\hat{D}^{\frac{1}{4}}}_t$ into a full resolution depth map $\mathbf{\hat{D}}^{\text{up}}_t$. We use the convex up-sampling~\cite{teed2020raft}, where we predict a mask of shape $(H/4) \times (W/4) \times (4 \times 4 \times 9)$. Each value in $\mathbf{\hat{D}}^{\text{up}}_t$ is calculated as the convex combination of the values of its $3 \times 3$ neighbors in $\mathbf{\hat{D}}^{\frac{1}{4}}_t$.

Finally, we use a spatial propagation network (SPN) to enhance the up-sampled depth map and get the final prediction $\mathbf{\hat{D}}_t$: $\mathbf{\hat{D}}_t = \operatorname{SPN}(\mathbf{\hat{D}}_t^{\text{up}}, \mathbf{\hat{F}^{\text{full}}})$.  SPN helps our model retain high-resolution details, especially at object boundaries. We use DySPN~\cite{lin2022dynamic} in our implementation, but our model can also work well with other SPNs or even without an SPN, because DDI plays the role of propagating the sparse depth information globally. 

At training time, we up-sample and enhance all intermediate depth maps for supervision. At test time, we only up-sample and enhance the last prediction.

\subsection{Loss Functions}

We supervise the final outputs $\{\mathbf{\hat{D}}_1, \cdots ,\mathbf{\hat{D}}_T\}$ at all iterations. We also supervise the up-sampled depth maps $\{\mathbf{\hat{D}}^{\text{up}}_1, \cdots ,\mathbf{\hat{D}^{\text{up}}}_T\}$ and the depth gradients $\{\mathbf{\hat{G}}_1, \cdots ,\mathbf{\hat{G}}_T\}$. Given a ground-truth depth map $\mathbf{D}$, we compute the ground-truth depth gradients $\mathbf{G}$ by down-sampling $\mathbf{D}$ and taking its finite difference.

We supervise $\mathbf{\hat{D}}$ and $\mathbf{\hat{D}}^{\text{up}}$ with a combination of an $L_1$ and an $L_2$ loss, and $\mathbf{\hat{G}}$ with an $L_1$ loss. We use a loss decay $\gamma=0.9$ and $\lambda=1.0$ in all experiments:

\begin{gather}
    \mathcal{L}_{\mathbf{D}} = \sum_{t=1}^T \gamma^{T-t} \left( \left\| \mathbf{\hat{D}}_t - \mathbf{D} \right\|_2^2 + 
    \left\| \mathbf{\hat{D}}_t - \mathbf{D} \right\|_1
    +
    \left\| \mathbf{\hat{D}^\text{up}}_t - \mathbf{D}\right\|_2^2 + 
    \left\| \mathbf{\hat{D}^\text{up}}_t - \mathbf{D} \right\|_1
    \right), 
    \\
    \mathcal{L}_{\mathbf{G}} = \sum_{t=1}^T \gamma^{T-t}  \left\| \mathbf{\hat{G}}_t - \mathbf{G} \right\|_1,
    \;
    \mathcal{L} = \mathcal{L}_{\mathbf{D}} + \lambda \cdot \mathcal{L}_{\mathbf{G}}.
\end{gather}

\section{Differentiable Depth Integrator (DDI)}
\label{sec:DDI}

In this section, we introduce our differentiable depth integrator (DDI), which converts the predicted depth gradients and the given sparse observations into a depth map. 
We define DDI as a function:
\begin{equation}
    \mathbf{\hat{D}} = \operatorname{DDI}(\mathbf{\hat{G}}, \mathbf{O}, \mathbf{M}).
\end{equation}

Let $H \times W$ be the depth map resolution. $\mathbf{\hat{G}} = \{ \mathbf{\hat{G}^x}, \mathbf{\hat{G}^y} \} \in \mathbb{R}^{2\times H \times W}$ is the predicted depth gradients along the $x$ and $y$ directions. $\mathbf{O} \in \mathbb{R}_+^{H \times W}$ is the sparse depth observations and $\mathbf{M} \in \{0,1\}^{H \times W}$ is its corresponding mask. $\mathbf{\hat{D}} \in \mathbb{R}_+^{H \times W}$ is the predicted depth map.

DDI is the key to achieving strong generalization and high accuracy, as it explicitly enforces the predicted depth map to be consistent with the sparse observations, introducing a strong inductive bias to the network. Moreover, DDI is differentiable, allowing us to train the network end-to-end.

\littlesection{Optimization Problem Definition.} DDI solves a 2D numerical integration problem, integrating depth gradients into depth with boundary conditions set by the sparse observations. We formulate depth integration as a linear least squares problem with two energy terms:
 
\begin{equation}
    \label{eq:optim_target}
    \mathbf{\hat{D}} = \argmin_{\mathbf{D}} \left( E_G (\mathbf{D}, \mathbf{\hat{G}}) + \alpha \cdot E_O (\mathbf{D}, \mathbf{O},\mathbf{M} ) \right),
\end{equation}
where $E_G(\cdot)$ corresponds to the gradient conditions, $E_O(\cdot)$ corresponds to the observation conditions, and $\alpha$ is a hyperparameter balancing the relative impact of the two terms. The performance of OGNI-DC is not sensitive to $\alpha$ and we use $\alpha=5.0$ for all experiments.

$E_G(\cdot)$ encourages the finite differences between neighboring pixels in $\mathbf{D}$ to be close to the predicted depth gradients $\mathbf{\hat{G}}$:
\begin{equation}
    E_G (\mathbf{D}, \mathbf{\hat{G}}) = \sum_{i=2}^W \sum_{j=1}^H \left( \mathbf{D}_{i,j} - \mathbf{D}_{i-1, j} - \mathbf{\hat{G}^x}_{i,j} \right) ^ 2 + \sum_{i=1}^W \sum_{j=2}^H \left( \mathbf{D}_{i,j} - \mathbf{D}_{i, j-1} - \mathbf{\hat{G}^y}_{i,j} \right) ^ 2.
\end{equation}

$E_O(\cdot)$ encourages the predicted depth values to be consistent with the sparse observations at valid locations: 

\begin{equation}
    E_O (\mathbf{D}, \mathbf{O},\mathbf{M}) =  \sum_{i=1}^W \sum_{j=1}^H \mathbf{M}_{i,j} \cdot (\mathbf{D}_{i,j} - \mathbf{O}_{i,j})^2.
\end{equation}

We model the observation conditions as energy terms rather than hard constraints for two reasons. 1) The optimization problem with soft constraints can be solved more efficiently. 2) The sparse observations can contain noise, such as the bleeding artifacts caused by blended foreground and background depth in KITTI~\cite{kittidc}. We solve this problem by predicting a confidence map for the sparse depth input, which allows DDI to ignore noisy sparse depth values. We omit it here for brevity. Please refer to the Appendix for details.

\littlesection{Conjugate Gradient Solver.} We rewrite \cref{eq:optim_target} in the matrix form of linear least squares:
\begin{equation}
\label{eq:least_square_matrix_form}
\renewcommand\arraystretch{1.3}
\overline{\hat{\mathbf{D}}} = \argmin_{\overline{\mathbf{D}}} 
\left\| 
\underbrace{\begin{pmatrix}
\mathbf{\overline{\Delta_x}} \\
\mathbf{\overline{\Delta_y}} \\
\operatorname{diag}\left(\sqrt{\alpha} \cdot \overline{\mathbf{M}} \right)  \\
\end{pmatrix}}_{\mathbf{A}}
\mathbf{\overline{D}} - 
\underbrace{\begin{pmatrix}
\overline{\mathbf{\hat{G}^x}} \\
\overline{\mathbf{\hat{G}^y}} \\
\sqrt{\alpha} \cdot \overline{\mathbf{M}}\cdot \overline{\mathbf{O}} \\
\end{pmatrix}}_{\mathbf{b}}
\right\| _2^2,
\end{equation}
where $\mathbf{\Delta_x}$ is the finite difference operator in the $x$ direction, \ie, $(\mathbf{\Delta_x} \circ \mathbf{D})_{i,j} = \mathbf{D}_{i,j} - \mathbf{D}_{i-1,j}$, and $\mathbf{\Delta_y}$ is defined accordingly. The bar ($\overline{\;\cdot\;}$) over a matrix or an operator means the flattened version of it. 

\cref{eq:least_square_matrix_form} has the closed-form solution of $\overline{\hat{\mathbf{D}}} = (\mathbf{A}^\intercal \mathbf{A})^{-1}\mathbf{A}^\intercal \mathbf{b}$. However, solving $\overline{\hat{\mathbf{D}}}$ by directly computing $(\mathbf{A}^\intercal \mathbf{A})^{-1}$ is intractable for high-resolution images, as $\mathbf{A}^\intercal \mathbf{A}$ is a giant matrix with shape $HW \times HW$. Therefore, we use the conjugate gradient method~\cite{hestenes1952methods} to solve it efficiently. Note that neither $\mathbf{A}$ nor $\mathbf{A}^\intercal \mathbf{A}$ needs to be explicitly stored, as we only need to implement its matrix multiplication with a vector. All operations are implemented in PyTorch~\cite{paszke2019pytorch} and naturally support GPU acceleration.

\littlesection{Backward Pass.} To make DDI differentiable, we have to compute the Jacobian matrix $\partial \mathbf{\hat{D}} / \partial \mathbf{\hat{G}}$. One option is to trace through the whole conjugate gradient process in the forward pass. However, the memory cost is intractable as the number of conjugate gradient steps is large. Instead, we compute $\partial \mathbf{\hat{D}} / \partial \mathbf{\hat{G}}$ by the chain rule:
\begin{gather}
\label{eq:backward_pass}
\renewcommand\arraystretch{1.4}
\begin{matrix}
    \partial \overline{\mathbf{\hat{D}}} \\ \hline
    \partial \overline{\mathbf{\hat{G}}}
\end{matrix} 
= 
\begin{matrix}
    \partial \overline{\mathbf{\hat{D}}} \\ \hline
    \partial \mathbf{b}
\end{matrix}
\cdot 
\begin{matrix}
    \partial \mathbf{b} \\ \hline
    \partial \overline{\mathbf{\hat{G}}}
\end{matrix}\;,
\\
\enspace
\renewcommand\arraystretch{1.4}
\begin{matrix}
    \partial \overline{\mathbf{\hat{D}}} \\ \hline
    \partial \mathbf{b}
\end{matrix} 
= 
(\mathbf{A}^\intercal \mathbf{A} )^{-1} \mathbf{A}^\intercal
,
\enspace
\begin{matrix}
    \partial \mathbf{b} \\ \hline
    \partial \overline{\mathbf{\hat{G}}}
\end{matrix}
\;
= 
\begin{pmatrix}
\mathbf{I}_{H(W-1)} & \mathbf{0} &  \mathbf{0} \\
\mathbf{0} & \mathbf{I}_{(H-1)W} &  \mathbf{0} \\
\end{pmatrix}^\intercal
,
\end{gather}
where $\mathbf{I}$ is the identity matrix. 
Note again we do not have to explicitly compute or store $\partial \overline{\mathbf{\hat{D}}} / \partial \mathbf{b}$, as we only need its matrix multiplication with a vector. This can be done effectively using the same conjugate gradient solver as in the forward pass. Details can be found in the Appendix. 

\littlesection{Initialization from Previous Solution.} 
Since we solve the optimization problem multiple times in each forward and backward pass, each time with slightly refined depth gradients, the solution from the previous round can be used as an initial guess to accelerate convergence. This reduces the overall DDI latency by up to 62.1\%. Please see \cref{tab:ablation} for detailed analysis of speed.

\section{Experiments}
\subsection{Implementation Details}
We implement OGNI-DC in PyTorch~\cite{paszke2019pytorch}. We use the AdamW~\cite{loshchilov2018decoupled} optimizer with an initial learning rate of 0.001. We use 5 GRU iterations for all experiments. On NYUv2~\cite{nyuv2}, we train the model on a single RTX 3090 GPU for 36 epochs with a batch size of 12, which takes about 3 days. On KITTI~\cite{kittidc}, we train the model on 8$\times$L40 GPUs for 100 epochs with a batch size of 40, which takes about 1 week. Following previous works~\cite{tang2019learning,zhao2021adaptive,wang2023lrru}, we average the predictions for the original and the horizontal-flipped inputs only for the KITTI online submissions.

Inspired by previous works~\cite{conti2023sparsity,long2023sparsedc}, to make the backbone more robust to depth density changes, we use a simple random masking technique on the sparse depth input: at training time, we randomly drop $0\sim 100 \%$ observed depth values for $50 \%$ training samples, and keep the other $50 \%$ untouched. 

\subsection{Datasets and Evaluation Metrics}
We evaluate our method on 4 commonly used datasets. To provide good coverage of both indoor and outdoor scenes, we use NYUv2~\cite{nyuv2} and VOID~\cite{void} for room environments, and KITTI~\cite{kittidc} and DDAD~\cite{ddad} for autonomous driving environments. Detailed descriptions of the datasets' statistics, split, and pre-processing can be found in the Appendix.

\littlesection{Evaluation Metrics.} We evaluate under the standard metrics: \textit{Root Mean Squared Error (RMSE)}, \textit{Mean Absolute Error (MAE)}, \textit{Root Mean Squared Error of Inverse Depth (iRMSE)}, \textit{Mean Absolute Error of Inverse Depth (iMAE)}, and \textit{Mean Absolute Relative Error (REL)}. Definitions can be found in \cite{khan2022comprehensive}.

\subsection{Zero-shot Generalization to VOID and DDAD}

\begin{figure}[t]
  \centering
  \includegraphics[width=\linewidth]{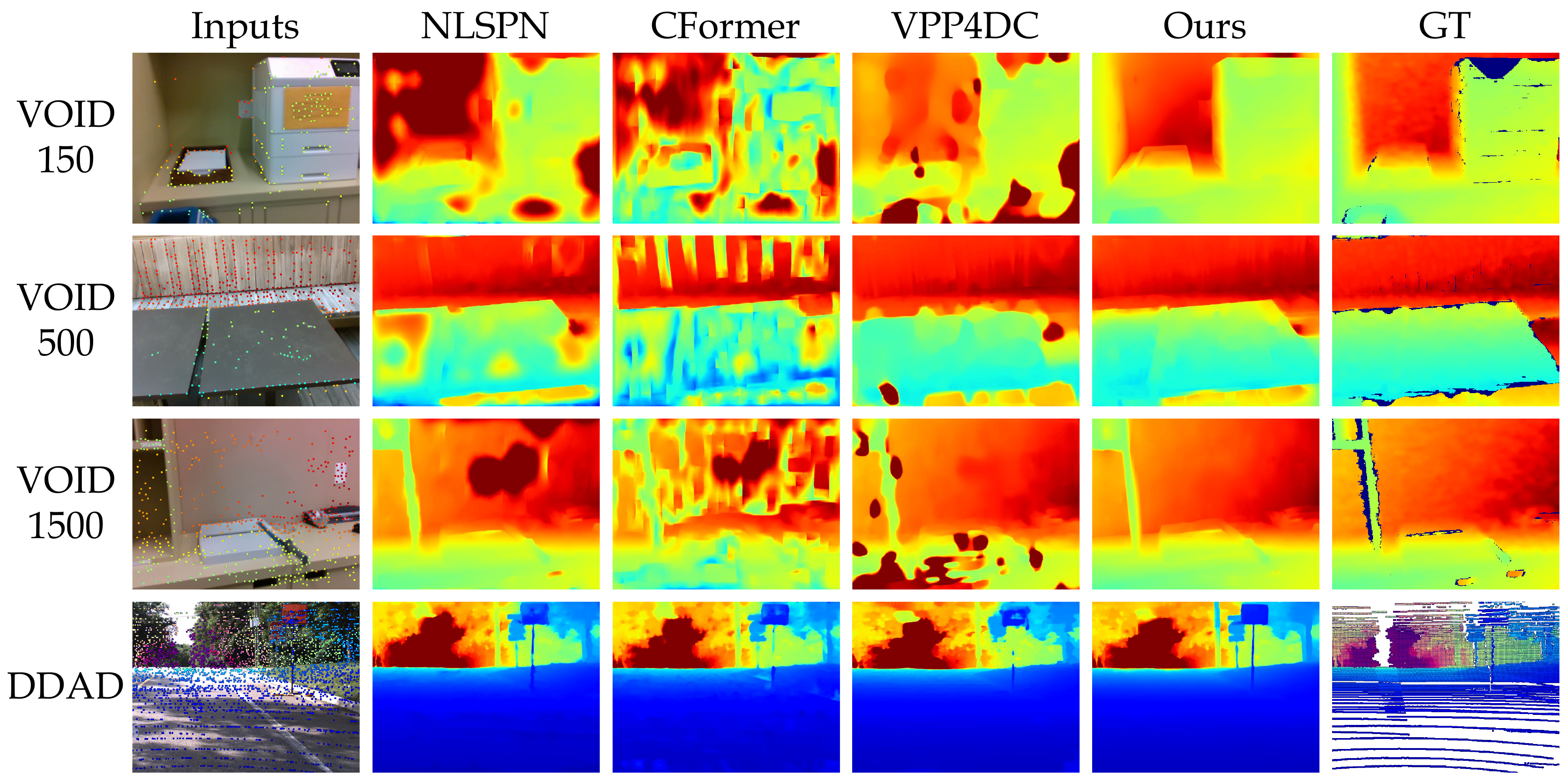}
  \caption{Qualitative comparisons of zero-shot generalization on the VOID~\cite{void} and the DDAD~\cite{ddad} datasets. The sparse depth observations are superimposed on the RGB images. Compared to baselines, our results are less noisy and sharper at boundaries.}
  \label{fig:void_ddad}
\end{figure}

\begin{table}[t]
\setlength{\tabcolsep}{2mm}
\centering
        \caption{Zero-shot generalization to VOID~\cite{void} and DDAD~\cite{ddad}. All metrics are in {[}m{]}.}
        \label{tab:void_ddad}
            \begin{tabular}{@{}lcccccccc@{}}
              \hline
              \noalign{\vskip 0.5mm}
  Train Datasets & \multicolumn{6}{c}{NYU} & \multicolumn{2}{c}{KITTI} \\
  
  \hline
  \noalign{\vskip 0.5mm}
  Test Datasets &
  \multicolumn{2}{c}{VOID1500~\cite{void}} &
  \multicolumn{2}{c}{VOID500~\cite{void}} &
  \multicolumn{2}{c}{VOID150~\cite{void}} &
  \multicolumn{2}{c}{DDAD~\cite{ddad}}
  \\ 
  Points/Density &
  \multicolumn{2}{c}{\numprint{1500}/0.5\%} &
  \multicolumn{2}{c}{\numprint{500}/0.15\%} &
  \multicolumn{2}{c}{\numprint{150}/0.05\%} &
  \multicolumn{2}{c}{\numprint{5000}/0.21\%}
  \\
  \cmidrule(lr){2-3} \cmidrule(lr){4-5} \cmidrule(lr){6-7} \cmidrule(lr){8-9}
  Methods &
  RMSE & MAE &
  RMSE & MAE &
  RMSE & MAE & RMSE & MAE
  
  \\
  \hline \hline  
  NLSPN~\cite{park2020non} & 0.737 & 0.298 & 0.802 & 0.381 & 0.963 & 0.492 & 11.646 & 4.621 \\
  SpAgNet~\cite{conti2023sparsity} & 0.706 & 0.244 & 0.752 & 0.326 & 0.866 & 0.408 & 18.247 & 9.130 \\
  CFormer~\cite{zhang2023completionformer} & 0.726 & 0.261 & 0.821 & 0.385 & 0.956 & 0.487 & 9.606 & 3.328 \\
  LRRU~\cite{wang2023lrru} & - & - & - & - & - & - & 9.164 & 2.738 \\
  VPP4DC~\cite{bartolomei2023revisiting} & 0.800 & 0.253 & 0.840 & 0.307 & 0.960 & 0.397 & 10.247 & 2.290 \\
  \hline
  Ours & \textBF{0.593} & \textBF{0.175} & \textBF{0.589} & \textBF{0.198} & \textBF{0.693} & \textBF{0.261} & \textBF{6.876} & \textBF{1.867} \\
  \hline
  
  \end{tabular}
\end{table}

The generalization to unseen environments is critical to the users of the depth completion systems. Following VPP4DC~\cite{bartolomei2023revisiting}, we test the \textit{zero-shot} generalization of our model in both indoor and outdoor environments. We train models on NYUv2~\cite{nyuv2} and KITTI~\cite{kittidc} and test them on VOID~\cite{void} and DDAD~\cite{ddad}, respectively. We compare against state-of-the-art baselines, including 3 models with the best in-domain performance~\cite{park2020non,zhang2023completionformer,wang2023lrru}, and VPP4DC~\cite{bartolomei2023revisiting} specially designed for cross-dataset generalization.

Results are shown in \cref{tab:void_ddad}. OGNI-DC outperforms baselines by a large margin on all metrics. On the most challenging VOID150~\cite{void} benchmark (about 0.05\% depth coverage), OGNI-DC improves MAE by 34.2\% (from 0.397 to 0.261) compared to the best-performing prior work VPP4DC~\cite{bartolomei2023revisiting}. On DDAD~\cite{ddad}, OGNI-DC improves RMSE by 25.0\% (from 9.164 to 6.876) compared to LRRU~\cite{wang2023lrru}. We visualize the depth predictions of different methods in \cref{fig:void_ddad}. Compared to baselines, our method generates sharp and accurate results on DDAD~\cite{ddad} and at all sparsity levels on VOID~\cite{void}.

\subsection{Robustness to Different Sparsity Levels}

\begin{table}[t]
\setlength{\tabcolsep}{2.1mm}
\centering
        \caption{Robustness to the number of Lidar lines on the KITTI~\cite{kittidc} validation set. All methods have a single model trained with 64 lines. All metrics are in {[}mm{]}.}
        \label{tab:kitti_sparse}
            \begin{tabular}{@{}lcccccccc@{}}
  \hline
  \noalign{\vskip 0.5mm}
  Lidar Scans &
  \multicolumn{2}{c}{64-Lines} &
  \multicolumn{2}{c}{32-Lines} &
  \multicolumn{2}{c}{16-Lines} &
  \multicolumn{2}{c}{8-Lines}
  \\ \cmidrule(l){2-3} \cmidrule(l){4-5} \cmidrule(l){6-7} \cmidrule(l){8-9}
  Methods &
  RMSE & MAE &
  RMSE & MAE &
  RMSE & MAE &
  RMSE & MAE 
  \\
  \hline \hline 
  NLSPN~\cite{park2020non} & 778.0 & 199.5 & 1217.2 & 367.5 & 1988.5 & 693.1 & 3234.9 & 1491.3 \\
  SpAgNet~\cite{conti2023sparsity} & 844.8 & 218.4 & 1164.2 & 339.2 & 1863.3 & 606.9 & 2691.3 & 1087.2 \\
  LRRU~\cite{wang2023lrru} & \textBF{729.5} & \textBF{188.8} & 1082.9 & 315.6 & 1978.5 & 700.5 & 3512.8 & 1624.0 \\
  CFormer~\cite{zhang2023completionformer} & 741.4 & 195.0 & 1241.8 & 384.9 & 2236.0 & 880.4 & 3638.8 & 1698.8 \\
  \hline 
  Ours & 749.8 & 192.9 & \textBF{1017.2} & \textBF{267.4} & \textBF{1661.8} & \textBF{451.9} & \textBF{2389.8} & \textBF{784.9} \\
  \hline
  \end{tabular}
\end{table}

\begin{table}[t]
\setlength{\tabcolsep}{2mm}
\centering
        \caption{Generalization to different sparsity levels on the NYUv2~\cite{nyuv2} test set. All methods have a single model trained with 500 points. The RMSE metric is in {[}m{]}. ``X+Mask'' are the baselines retrained by randomly masking out $0\sim 100 \%$ of the sparse depth observations during training.}
        \label{tab:nyu_sparse}
            \begin{tabular}{@{}lcccccccc@{}}
  \hline
  \noalign{\vskip 0.5mm}
  Samples & 
  \multicolumn{2}{c}{20000} &
  \multicolumn{2}{c}{5000} &
  \multicolumn{2}{c}{1000} &
  \multicolumn{2}{c}{500} 
  \\ \cmidrule(lr){2-3} \cmidrule(r){4-5} \cmidrule(lr){6-7} \cmidrule(lr){8-9}
  Methods & 
  RMSE & REL &
  RMSE & REL &
  RMSE & REL &
  RMSE & REL \\
  \hline \hline 
  NLSPN~\cite{park2020non} & 0.034 & 0.004 & \textBF{0.042} & \textBF{0.005} & 0.071 & \textBF{0.009} & 0.092 & \textBF{0.012} \\
  SpAgNet~\cite{conti2023sparsity}  & - & - & - & - & - & - & 0.114 & 0.015 \\
  CFormer~\cite{zhang2023completionformer} & 0.203 & 0.046 & 0.045 & \textBF{0.005} & \textBF{0.070} & \textBF{0.009} & 0.090 & \textBF{0.012} \\
  NLSPN+Mask & 0.796 & 0.133 & 0.086 & 0.013 & 0.082 & 0.011 & 0.103 & 0.014 \\
  CFormer+Mask & 0.161 & 0.022 & 0.050 & 0.007 & 0.075 & 0.010 & 0.096 & 0.013 \\
  \hline 
  Ours & \textBF{0.028} & \textBF{0.003} & \textBF{0.042} & \textBF{0.005} & \textBF{0.070} & \textBF{0.009} & \textBF{0.089} & \textBF{0.012} \\
  \hline

  \noalign{\vskip 2mm}

  \hline
  \noalign{\vskip 0.5mm}
  Samples &
  \multicolumn{2}{c}{200} &
  \multicolumn{2}{c}{100} &
  \multicolumn{2}{c}{50} &
  \multicolumn{2}{c}{5}
  \\ \cmidrule(lr){2-3} \cmidrule(lr){4-5} \cmidrule(lr){6-7} \cmidrule(lr){8-9}
  Methods & 
  RMSE & REL & 
  RMSE & REL &
  RMSE & REL &
  RMSE & REL \\
  \hline \hline 
  NLSPN~\cite{park2020non} & 0.136 & 0.019 & 0.245 & 0.037 & 0.431 & 0.081 & 1.043 & 0.262 \\
  SpAgNet~\cite{conti2023sparsity} & 0.155 & 0.024 & 0.209 & 0.038 & 0.272 & 0.058 & \textBF{0.467} & 0.131 \\
  CFormer~\cite{zhang2023completionformer} & 0.141 & 0.021 & 0.429 & 0.092 & 0.707 & 0.181 & 1.141 & 0.307 \\
  NLSPN+Mask & 0.142 & 0.022 & 0.183  & 0.031 & 0.241 & 0.046 & 0.728 & 0.175 \\
  CFormer+Mask & 0.135 & 0.021 & 0.174 & 0.030 & 0.227 & 0.044 & 0.487 & \textBF{0.129} \\
  \hline 
  Ours & \textBF{0.124} & \textBF{0.018} & \textBF{0.157} & \textBF{0.025} & \textBF{0.207} & \textBF{0.038} & 0.633 & 0.171 \\
  \hline
  \end{tabular}
  \vspace{-3mm}
\end{table}

Robustness to different sparsity levels is also important for downstream applications. Previous works~\cite{imran2021depth,zhang2023completionformer} \textit{retrain} different models at different sparsity levels. We find this setting impractical, as the depth density cannot be known \textit{a priori} in real testing environments. For example, when the sparse observations come from a SLAM system~\cite{mur2017orb}, the sparsity level can vary drastically from frame to frame as various numbers of key points are being tracked. Therefore, we want \textit{one model} to perform well across all sparsity levels.

We adopt the experiment setup from SpAgNet~\cite{conti2023sparsity}, where all models are trained under the standard setting (500 points on NYUv2~\cite{nyuv2}, 64-Lines Lidar on KITTI~\cite{kittidc}), and tested with different sparsity levels. We compare against several baselines, including SpAgNet~\cite{conti2023sparsity} which is specialized for sparsity generalization. We build two stronger baselines, where we retrain the NLSPN~\cite{park2020non} and CFormer~\cite{zhang2023completionformer} with the same random masking we use, which we denote as ``X+Mask''. 

On KITTI~\cite{kittidc}, we sub-sample the Lidar points following \cite{imran2021depth} to 32, 16, and 8 lines. On NYUv2~\cite{nyuv2}, we test across a wide range of $5\sim\numprint{20000}$ randomly sampled sparse points. 
Results are shown in \cref{tab:kitti_sparse} and \cref{tab:nyu_sparse}. 

On KITTI~\cite{kittidc}, OGNI-DC achives comparable performance as baselines on 64-Lines input and outperforms baselines by a large margin on sparser inputs, reducing MAE by 25.5\% compared to SpAgNet~\cite{conti2023sparsity} (451.9 versus 606.9) on 16-Lines input. 

On NYUv2~\cite{nyuv2}, our method works better than baselines under almost all sparsity levels. 
Interestingly, our method can also work well with more than 500 points, although it has never seen these cases during training. The generalization of CFormer~\cite{zhang2023completionformer} and NLSPN~\cite{park2020non} improves with random masking, but is still worse than ours. 
Under the extremely sparse case (5 points), our method still works better than NLSPN~\cite{park2020non} and CFormer~\cite{zhang2023completionformer}, but slightly worse than SpAgNet~\cite{conti2023sparsity}. That's probably because the estimation errors on depth gradients can accumulate in the integration process in large regions without any depth observations. However, we don't find this to be a significant limitation of our system, since these cases are very unlikely in real-world scenarios (\eg, ORB-SLAM~\cite{mur2015orb} cannot add new key-frames when it tracks fewer than 50 key-points). 

\subsection{In-domain Performance on NYUv2 and KITTI}

\begin{table}[t]
\setlength{\tabcolsep}{2.3mm}
\centering
        \caption{Comparison with prior works on the NYUv2~\cite{nyuv2} and KITTI~\cite{kittidc} datasets. We mark the best method in \textbf{bold} and the second-best method with an \underline{underline}.}
        \label{tab:NYUv2_KITTI}
            \begin{tabular}{@{}lccccccc@{}}
            \hline
            \noalign{\vskip 0.5mm}            
            Datasets & \multicolumn{2}{c}{NYUv2~\cite{nyuv2}} &  & \multicolumn{4}{c}{KITTI~\cite{kittidc}} \\ \hline
            \noalign{\vskip 0.5mm}            Methods & \begin{tabular}{@{}c@{}}{RMSE} \\ {[}m{]} \end{tabular} & REL &  & \begin{tabular}{@{}c@{}}{iRMSE} \\ {[}1/km{]} \end{tabular} & \begin{tabular}{@{}c@{}}{iMAE} \\ {[}1/km{]} \end{tabular} & \begin{tabular}{@{}c@{}}{RMSE} \\ {[}mm{]} \end{tabular} & \begin{tabular}{@{}c@{}}{MAE} \\ {[}mm{]} \end{tabular} \\ \hline
            CSPN~\cite{cheng2019cspn} & 0.117 & 0.016 &  & 2.93 & 1.15 & 1019.64 & 279.46 \\
            DeepLiDAR~\cite{qiu2019deeplidar} & 0.115 & 0.022 &  & 2.56 & 1.15 & 758.38 & 226.50 \\
            GuideNet~\cite{tang2019learning} & 0.101 & 0.015 &  & 2.25 & 0.99 & 736.24 & 218.83 \\
            NLSPN~\cite{park2020non} & 0.092 & \underline{0.012} &  & 1.99 & 0.84 & 741.68 & 199.59 \\
            ACMNet~\cite{zhao2021adaptive} & 0.105 & 0.015 &  & 2.08 & 0.90 & 744.91 & 206.09 \\
            RigNet~\cite{yan2021rignet} & 0.090 & \underline{0.012} &  & 2.08 & 0.90 & 712.66 & 203.25 \\
            DySPN~\cite{lin2022dynamic} & 0.090 & \underline{0.012} &  & 1.88 & 0.82 & 709.12 & 192.71 \\
            LRRU~\cite{wang2023lrru} & 0.091 & \textBF{0.011} &  & 1.87 & 0.81 & \textBF{696.51} & 189.96 \\
            BEV@DC~\cite{zhou2023bev} & \underline{0.089} & \underline{0.012} &  & \underline{1.83} & 0.82 & \underline{697.44} & 189.44 \\ CFormer~\cite{zhang2023completionformer} ($L_1$) & - & - &  & 1.89 & \underline{0.80} & 764.87 & \underline{183.88} \\       CFormer~\cite{zhang2023completionformer} ($L_1 + L_2$) & 0.090 & \underline{0.012} &  & 2.01 & 0.88 & 708.87 & 203.45 \\
            \hline
            Ours ($L_1$) & - & - &  & \textBF{1.81} & \textBF{0.79} & 747.64 & \textBF{182.29} \\
            Ours ($L_1 + L_2$) & \textBF{0.087} & \textBF{0.011} &  & 1.86 & 0.83 & 708.38 & 193.20 \\ \hline
            \end{tabular}
\end{table}

\begin{figure}[t]
  \centering
  \includegraphics[width=\linewidth]{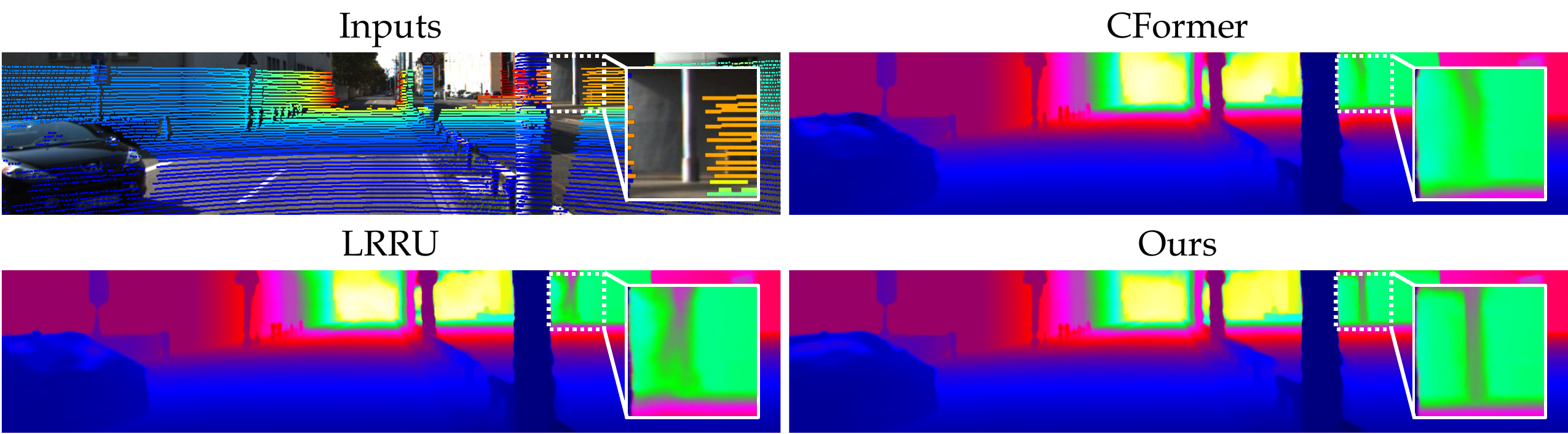}
  \caption{Qualitative comparison with LRRU~\cite{wang2023lrru} and CFormer~\cite{zhang2023completionformer} on the KITTI~\cite{kittidc} test split. Our method reconstructs the telephone pole better than the baselines.}
  \label{fig:kitti}
\end{figure}

We compare the in-domain performance of OGNI-DC with other state-of-the-art methods on the standard NYUv2~\cite{nyuv2} and KITTI~\cite{kittidc} benchmarks.

Quantitative results are shown in \cref{tab:NYUv2_KITTI}. On the NYUv2~\cite{nyuv2} dataset, our model achieves RMSE of 0.087 and REL of 0.011, the best among all previous methods. On the KITTI~\cite{kittidc} dataset, following CFormer~\cite{zhang2023completionformer}, we train two models, one with only $L_1$ loss and the other with combined $L_1+L_2$ loss. Our model trained with $L_1$ loss archives the state-of-the-art MAE of 182.29, iRMSE of 1.81, and iMAE of 0.79. Our model trained with combined $L_1+L_2$ loss achieves a better RMSE of 708.38, which is close to the prior best-performing model. Note that in \cref{tab:NYUv2_KITTI} we disable masking to achieve the best in-domain performance and for a fair comparison with previous methods. The difference caused by masking is marginal: our model with masking still archives the state-of-the-art performance on the NYUv2~\cite{nyuv2} with  $\text{RMSE}=0.089$ (see \cref{tab:nyu_sparse}, 500 samples). 

Qualitative comparisons on the KITTI~\cite{kittidc} test split are shown in \cref{fig:kitti}. Our method can reconstruct the thin telephone pole, on which both baselines fail.

\subsection{Ablation Studies}
To study the effects of the key designs in OGNI-DC, we conduct ablation studies on the NYUv2~\cite{nyuv2} and the KITTI~\cite{kittidc} validation set. On KITTI, following CFormer~\cite{zhang2023completionformer}, we train models on 10,000 images. Results are shown in \cref{tab:ablation}. 

\begin{table}[t]

\setlength{\tabcolsep}{0.85mm}
\renewcommand{\arraystretch}{1}
\centering
\caption{We collect statistics with PyTorch Profiler on a L40 GPU under the KITTI resolution ($240\times1216$). All metrics are in mm.}
        \label{tab:ablation}
            \begin{tabular}{@{}lcccccccccc@{}}
              \hline
              \noalign{\vskip 1.0mm}  
  Metrics &
  \multicolumn{2}{c}{NYUv2~\cite{nyuv2}} &
  \multicolumn{2}{c}{KITTI~\cite{kittidc}} &
  \multicolumn{4}{c}{Inference Time (ms)}  & \multirow{2}{*}{\begin{tabular}{@{}c@{}}{Mem} \\ {(GB)} \end{tabular}}
  \\
  \cmidrule(lr){2-3} \cmidrule(lr){4-5} \cmidrule(lr){6-9}
  Methods & 
  RMSE & MAE &
  RMSE & MAE & 
  \textBF{Total} & Backbone & GRU & DDI 
  \\
  \noalign{\vskip 0.5mm}
  \hline \hline  
  CFormer+DySPN & 123.6 & 43.2 & 825.1 & 208.8 &  268.0 &  268.0 & 0.0 & 0.0 & 5.4 \\
  \hline
  No DDI & 128.6 & 44.9 & 824.0 & 207.5 & 298.7 & 256.5 & 42.2 & 0.0 & 5.9 \\
  DDI zeros init & \textBF{112.2} & 38.0 & 813.7 & \textBF{205.4} & 646.5 & 256.5 & 50.5 & 339.5 & 6.0  \\
  DDI pre-filled init & \textBF{112.2} & 38.0 & 813.7 & \textBF{205.4} & 601.9 & 256.5 & 50.5 & 294.9 & 6.0  \\
  \hline
  1 GRU iteration & 114.0 & 39.9 & 820.1 & 208.8 & 317.6 & 256.5 & 9.8 & 51.3 & 5.2 \\
  3 GRU iterations & 112.4 & 38.2 & 818.6 & 207.8 & 379.7 & 256.5 & 28.8 & 94.4 & 5.6 \\
  7 GRU iterations & 112.3 & 38.0 & \textBF{811.3} & 205.5 & 484.2 & 256.5 & 68.0 & 159.7  & 6.3 \\
  \hline
  ConvRNN & 112.7 & 38.1 & 817.7 & 205.8 & 408.7 & 256.5 & 34.0 & 118.2 & 5.6  \\
  Cascaded & 112.5 &  \textBF{37.9} & 814.4 & 206.0 & 444.7 & 256.5 & 50.5 & 137.7 & 5.9 \\
  \hline
  Ours & \textBF{112.2} & 38.0 & 813.7 & \textBF{205.4} & 435.8 & 256.5 & 50.5 & 128.8 & 5.9  \\

  \hline
  \noalign{\vskip -5mm}
  \end{tabular}
\end{table}

\littlesection{CFormer$+$DySPN.} Our model works significantly better than this baseline, improving the RMSE from 123.6mm to 112.2mm on NYU. Our model requires slightly more computation than the baseline: the FPS drops by 38\% and the memory usage increases by 10\%. We believe our model's state-of-the-art accuracy and generalization is worth this trade-off in computation.

\littlesection{DDI.} We train a model without DDI, where the ConvGRU directly generates updates on the depth map. Comparing (No DDI) to ours, DDI significantly improves MAE from 44.9mm to 38.0mm on NYU. We also examine the convergence speed of DDI using different initialization strategies. Our init. from current solution (\cref{sec:DDI}) effectively reduce the DDI latency by 62.1\% and 56.3\% compared to init. from zeros and using the heuristic-based pre-filled depth as in LRRU~\cite{wang2023lrru}. 

\littlesection{Iterative Refinement.} To prove the effectiveness of iterative refinement, we unroll the ConvGRU for $1\sim7$ times. On NYU, 5 iterations improve MAE from 39.9mm to 38.0mm compared to 1 iteration. Unrolling 7 iterations provides no further improvements. Therefore, we use 5 iterations as a balance between performance and speed. See \cref{fig:iter_refine} for qualitative effect of the iterative refinement.

\littlesection{ConvGRU} To show the advantage of ConvGRU over other iterative designs, we compare against ConvRNN and a ConvGRU without weighting tying among iterations (Cascaded). ConvGRU has higher capacity with the gated mechanism and therefore achieves better performance than ConvRNN. Ours achieves similar performance with fewer parameters compared to Cascaded.

\section*{Acknowledgments}
This work was primarily supported by the Intelligence Advanced Research Projects Activity (IARPA) via Department of Interior/ Interior Business Center (DOI/IBC) contract number 140D0423C0075. The U.S. Government is authorized to reproduce and distribute reprints for Governmental purposes notwithstanding any copyright annotation thereon. Disclaimer: The views and conclusions contained herein are those of the authors and should not be interpreted as necessarily representing the official policies or endorsements, either expressed or implied, of IARPA, DOI/IBC, or the U.S. Government.

% ---- Bibliography ----
%
% BibTeX users should specify bibliography style 'splncs04'.
% References will then be sorted and formatted in the correct style.
%
\bibliographystyle{splncs04}
\bibliography{main}
\newpage

\begin{flushleft}
\section*{Appendix}
\end{flushleft}

\setcounter{section}{0}
\renewcommand{\thesection}{\Alph{section}}
\setcounter{table}{0}
\renewcommand{\thetable}{\alph{table}}
\setcounter{figure}{0}
\renewcommand{\thefigure}{\alph{figure}}

\section{More Ablation Studies on Generalization}

\begin{table}[ht]
\setlength{\tabcolsep}{0.8mm}
\renewcommand{\arraystretch}{1.15}
\centering
        \caption{We conduct Ablation studies in the in-domain (NYUv2, 500 points), sparser inputs (NYUv2, 100 points), denser inputs (NYUv2, \numprint{10000} points), and cross-dataset (VOID, 500 points) scenarios. All metrics are in {[}mm{]}.}
        \label{tab:ablation_full}
            \begin{tabular}{@{}llcccccccc@{}}
              \hline
              \noalign{\vskip 0.5mm}  
  \noalign{\vskip 0.5mm}
  \multicolumn{2}{@{}l}{Test Datasets} &
  \multicolumn{2}{c}{NYUv2~\cite{nyuv2}} &
  \multicolumn{2}{c}{NYUv2~\cite{nyuv2}} &
  \multicolumn{2}{c}{NYUv2~\cite{nyuv2}} &
  \multicolumn{2}{c}{VOID~\cite{void}}
  \\ 
  \multicolumn{2}{@{}l}{Points} &
  \multicolumn{2}{c}{\numprint{500}} &
  \multicolumn{2}{c}{\numprint{100}} &
  \multicolumn{2}{c}{\numprint{10000}} &
  \multicolumn{2}{c}{\numprint{500}}
  \\
  \cmidrule(lr){3-4} \cmidrule(lr){5-6} \cmidrule(lr){7-8} \cmidrule(lr){9-10}
  \multicolumn{2}{@{}l}{Methods} &
  RMSE & MAE &
  RMSE & MAE &
  RMSE & MAE & RMSE & MAE
  
  \\
  \hline \hline  
  \noalign{\vskip 0.5mm}
  (a) & Without DDI & 128.6 & 44.9 & 199.0 & 87.9 & 73.5 & 22.6 & 620.5 & 237.9 \\
  \textBF{Ours} & With DDI & \textBF{112.2} & \textBF{38.0} & \textBF{171.5} & \textBF{72.7} & \textBF{53.0} & \textBF{13.9} & \textBF{605.0} & \textBF{229.0}  \\
  \hline
  (b) & 1 GRU iteration & 114.0 & 39.9 & 171.7 & 73.0 & 54.0 & 14.3 & 634.6 & 241.5 \\
  (c) & 3 GRU iterations & 112.4 & 38.2 & 171.6 & 73.0 & 54.1 & 14.3 & 641.5 & 247.1 \\
  \textBF{Ours} & 5 GRU iterations & \textBF{112.2} & \textBF{38.0} & \textBF{171.5} & 72.7 & 53.0 & 13.9 & \textBF{605.0} & \textBF{229.0} \\
  (d) & 7 GRU iterations & 112.3 & \textBF{38.0} & 172.5 & \textBF{71.8} & \textBF{52.0} & \textBF{13.6} & 
  649.2 & 257.5
  \\
  \hline
  (e) & RGB backbone & 121.3 & 43.1 & 176.9 & 79.4 & 67.8 & 18.1 & 659.9 & 273.9 \\
  (f) & RGBD (w/o masking) & \textBF{110.1} & \textBF{36.9} & 292.3 & 149.1 & 59.1 & 15.1 & 792.8 & 371.6 \\
  \textBF{Ours} & RGBD (w/ masking) & 112.2 & 38.0 & \textBF{171.5} & \textBF{72.7} & \textBF{53.0} & \textBF{13.9} & \textBF{605.0} & \textBF{229.0} \\
  \hline
  (g) & Without SPN & 113.0 & 38.3 & \textBF{170.9} & \textBF{71.3} & 68.1 & 21.3 & 662.8 & 265.2 \\
  (h) & With NLSPN~\cite{park2020non} & 112.8 & 38.5 & 179.0 & 76.2 & 64.1 & 20.5 & 693.2 &  286.0 \\
  \textBF{Ours} & With DySPN~\cite{lin2022dynamic} & \textBF{112.2} & \textBF{38.0} & 171.5 & 72.7 & \textBF{53.0} & \textBF{13.9} & \textBF{605.0} & \textBF{229.0} \\
  \hline
  (i) &  Without $\mathbf{\hat{D}^{\text{up}}}$ loss & 113.2 & 38.6 & 178.7 & 77.0 & 55.7 & 14.4 & 643.2 & 252.8 \\
  (j) & Without $\mathbf{\hat{G}}$ loss & 112.5 & 38.1 & \textBF{171.5} & \textBF{71.7} & 54.1 & 14.0 & 634.0 & 252.2 \\
  \textBF{Ours} & With both losses & \textBF{112.2} & \textBF{38.0} & \textBF{171.5} & 72.7 & \textBF{53.0} & \textBF{13.9} & \textBF{605.0} & \textBF{229.0} \\

  \hline
  \end{tabular}
\end{table}

To better understand how the generalization of OGNI-DC is affected by different components in our depth completion pipeline, we conduct ablation studies by varying the sparsity levels and datasets. Specifically, we test models on the NYUv2~\cite{nyuv2} validation set with sparser inputs (randomly sampled 100 points) and denser inputs (randomly sampled \numprint{10000} points). We further test the cross-dataset generalization by constructing a validation set for the VOID~\cite{void} dataset by randomly choosing 800 images from the VOID500~\cite{void} training set. Finally, for easier reference, we copy the numbers from the main paper for the in-domain cases (NYUv2~\cite{nyuv2}, randomly sampled 500 points). Results are shown in \cref{tab:ablation_full}.

\littlesection{DDI.} We train a baseline model without DDI, where the ConvGRU directly generates updates on the depth map. Comparing (a) to ours, our model with DDI outperforms the baseline model by large margins under all settings, proving that DDI is the key to achieving both in-domain accuracy and strong generalization.

\littlesection{Iterative Refinement.} To prove the effectiveness of iterative refinement, we train different models where the ConvGRU are unrolled $1\sim7$ times. Compared to 1 or 3 iterations, 5 iterations consistently improve the performance under all settings. For example, when tested on VOID, 5 iterations improve MAE from 634.6mm to 605.0mm compared to 1 iteration. The benefits of further unrolling 7 iterations are not clear, as 7 iterations achieve slightly better results in the NYUv2 \numprint{10000} points case and worse results when tested on VOID. Therefore, we use 5 iterations as a good balance between performance and speed.

\littlesection{Backbone Inputs.} We test against the baseline where we input only the RGB image to the backbone ((e), RGB backbone). We further test against a baseline where we input both the image and the sparse depth to the backbone, but do not randomly mask out the sparse depth map during training ((f), RGBD (w/o masking)). Our full model consistently outperforms the RGB baseline, proving that sparse depth inputs are helpful for depth gradients prediction. Compared to ours, the model without masking archives better in-domain performance, but has worse generalization overall. In conclusion, RGBD w/masking provides the best balance between in-domain performance and generalization.

\littlesection{SPN.} We ablate the effect of the SPN layer in our model. Comparing $\text{(g)}\sim\text{(h)}$, the model without an SPN works best under 100 samples on NYUv2, whereas the model with DySPN~\cite{lin2022dynamic} works best for 500 points, \numprint{10000} points, and on VOID. The model with NLSPN~\cite{park2020non} works worse in all cases. In conclusion, our model works best with DySPN~\cite{lin2022dynamic}, but can also work well without an SPN.

\littlesection{Auxiliary Losses.} Comparing $\text{(i)}\sim\text{(j)}$ to ours, supervising the up-sampled depth $\mathbf{\hat{D}^{\text{up}}}$ and the depth gradients $\mathbf{\hat{G}}$ both contribute to better performance under all test setting. The contribution of the supervision on $\mathbf{\hat{D}^{\text{up}}}$ is more significant. That's probably because supervising the output of the SPN layer is not enough for regularizing its input, and therefore intermediate supervisions are necessary.

\section{KITTI Sparsity Genealization with Retrained Models}

To analyze our model's performance more thoroughly on sparser inputs, we run experiments under the setup of CFormer~\cite{zhang2023completionformer}, where all models are \textit{retrained} on sub-sampled lidar lines and tested under the same sparsity. To be consistent with CFormer~\cite{zhang2023completionformer}, instead of using the entire training set, we use the file list they provide, where they randomly sample \numprint{10000} training images. We test under the sparsity of 8, 16, 32, and 64 lines. We don't test even sparser inputs because no autonomous driving vehicles are equipped with Lidar sparser than 8 lines. 

Results are shown in \cref{tab:retrain}. The 64-Lines and 16-Lines numbers are copied from CFormer~\cite{zhang2023completionformer}. The 32-Lines and 8-Lines numbers are reproduced by ourselves with their official training code. Our model still consistently outperforms all baselines under all sparsity levels in this retrain setting.

\begin{table}[ht]
\setlength{\tabcolsep}{2.2mm}
\centering
        \caption{Robustness to the number of Lidar lines on the KITTI~\cite{kittidc} validation dataset. All methods are \textit{retrained} under the corresponding sparsity levels with \numprint{10000} images. All metrics are in {[}mm{]}. Our model consistently outperforms baselines.}
        \label{tab:retrain}
            \begin{tabular}{@{}lcccccccc@{}}
  \hline
  \noalign{\vskip 0.5mm}
  Lidar Scans &
  \multicolumn{2}{c}{64-Lines} &
  \multicolumn{2}{c}{32-Lines} &
  \multicolumn{2}{c}{16-Lines} &
  \multicolumn{2}{c}{8-Lines}
  \\ \cmidrule(l){2-3} \cmidrule(l){4-5} \cmidrule(l){6-7} \cmidrule(l){8-9}
  Methods &
  RMSE & MAE &
  RMSE & MAE &
  RMSE & MAE &
  RMSE & MAE 
  \\
  \hline \hline 
  NLSPN~\cite{park2020non} & 889.4 & 238.8 & 1052.2 & 285.0 & 1288.9 & 377.2 & 1584.0 & 501.9 \\
  DySPN~\cite{lin2022dynamic} & 878.5 & 228.6 & - & - & 1274.8 & 366.4 & - & - \\
  CFormer~\cite{zhang2023completionformer} & 848.7 & 215.9 & 994.8 & 265.8 & 1218.6 & 337.4 & 1513.2 & 457.7 \\
  \hline 
  Ours & \textBF{813.7} & \textBF{205.4} & \textBF{967.9} & \textBF{252.3} & \textBF{1196.7} & \textBF{324.3} & \textBF{1510.6} & \textBF{444.0} \\
  \hline
  \end{tabular}
\end{table}

\section{Details of the Differentiable Depth Integrator (DDI)}
\subsection{Backward Pass Details}

Let $\mathcal{L}$ be the loss of the network. The input to the $\operatorname{backward}$ function is its gradient on $\hat{\mathbf{D}}$, \ie, $\partial\mathcal{L}/\partial\hat{\mathbf{D}}$. We want to compute $\partial\mathcal{L}/\partial\hat{\mathbf{G}}$. Recall from the paper that we have:

\begin{gather}
\label{eq:backward_pass_conf}
\renewcommand\arraystretch{1.4}
\begin{matrix}
    \partial \overline{\mathbf{\hat{D}}} \\ \hline
    \partial \mathbf{b}
\end{matrix} 
= 
(\mathbf{A}^\intercal \mathbf{A} )^{-1} \mathbf{A}^\intercal
,
\enspace
\begin{matrix}
    \partial \mathbf{b} \\ \hline
    \partial \overline{\mathbf{\hat{G}}}
\end{matrix}
\;
=
\begin{pmatrix}
\mathbf{I}_{H(W-1)} & \mathbf{0} &  \mathbf{0} \\
\mathbf{0} & \mathbf{I}_{(H-1)W} &  \mathbf{0} \\
\end{pmatrix}^\intercal
.
\end{gather}

Therefore,
\begin{gather}
\renewcommand\arraystretch{1.4}
\begin{matrix}
    \partial 
    \mathcal{L}
    \\ \hline
    \partial \overline{\mathbf{\hat{G}}}
\end{matrix} 
= 
\begin{matrix}
    \partial \mathcal{L} 
    \\ \hline
    \partial \overline{\mathbf{\hat{D}}}
\end{matrix} 
\cdot
\begin{matrix}
    \partial \overline{\mathbf{\hat{D}}} \\ \hline
    \partial \mathbf{b}
\end{matrix}
\cdot 
\begin{matrix}
    \partial \mathbf{b} \\ \hline
    \partial \overline{\mathbf{\hat{G}}}
\end{matrix} 
\\
\renewcommand\arraystretch{1.4}
= 
\left(
    (\mathbf{A}^\intercal \mathbf{A} )^{-1}
    \cdot 
    \left(
    \begin{matrix}
    \partial \mathcal{L} 
    \\ \hline
    \partial \overline{\mathbf{\hat{D}}}
    \end{matrix}
    \right)^\intercal
\right)^\intercal
\cdot
\mathbf{A}^\intercal
\cdot
\begin{pmatrix}
\mathbf{I}_{H(W-1)} & \mathbf{0} &  \mathbf{0} \\
\mathbf{0} & \mathbf{I}_{(H-1)W} &  \mathbf{0} \\
\end{pmatrix}^\intercal.
\end{gather}

The first part can be computed using the same conjugate gradient solver as in the forward pass. Since $\mathbf{A}^\intercal$ and $\partial\mathbf{b} / \partial\overline{\hat{\mathbf{G}}}$ are sparse, the rest of the matrix multiplications can be done efficiently.

\subsection{Speed Optimizations}

\littlesection{Stopping Conditions.}
We stop the conjugate gradient solver when the optimal solution is found, \ie, when the relative residual is smaller than $1e-5$ or the residual has no improvement more than $1\%$ for $10$ steps. This avoids unnecessary conjugate gradient steps compared to optimizing for a fixed number of steps.

\littlesection{Initialization from Current Solution.} We assume the refinements on depth gradients to be small for each iteration, therefore the correct solution can be used as an initial guess for the next round. We store $\hat{\mathbf{D}}_t$ as a dummy variable and use it to be the initial value of $\hat{\mathbf{D}}_{t+1}$ in the conjugate gradient iterations. We do a similar thing for the backward pass by storing $
    (\mathbf{A}^\intercal \mathbf{A} )^{-1}
    \cdot 
    \partial \mathcal{L} / \partial \overline{\mathbf{\hat{D}}}$,
but this time using the result from the ($t+1$)-th step to initialize the $t$-th step.

\subsection{Confidence on Sparse Depth Observations}

\begin{figure}[t]
  \centering
  \includegraphics[width=\linewidth]{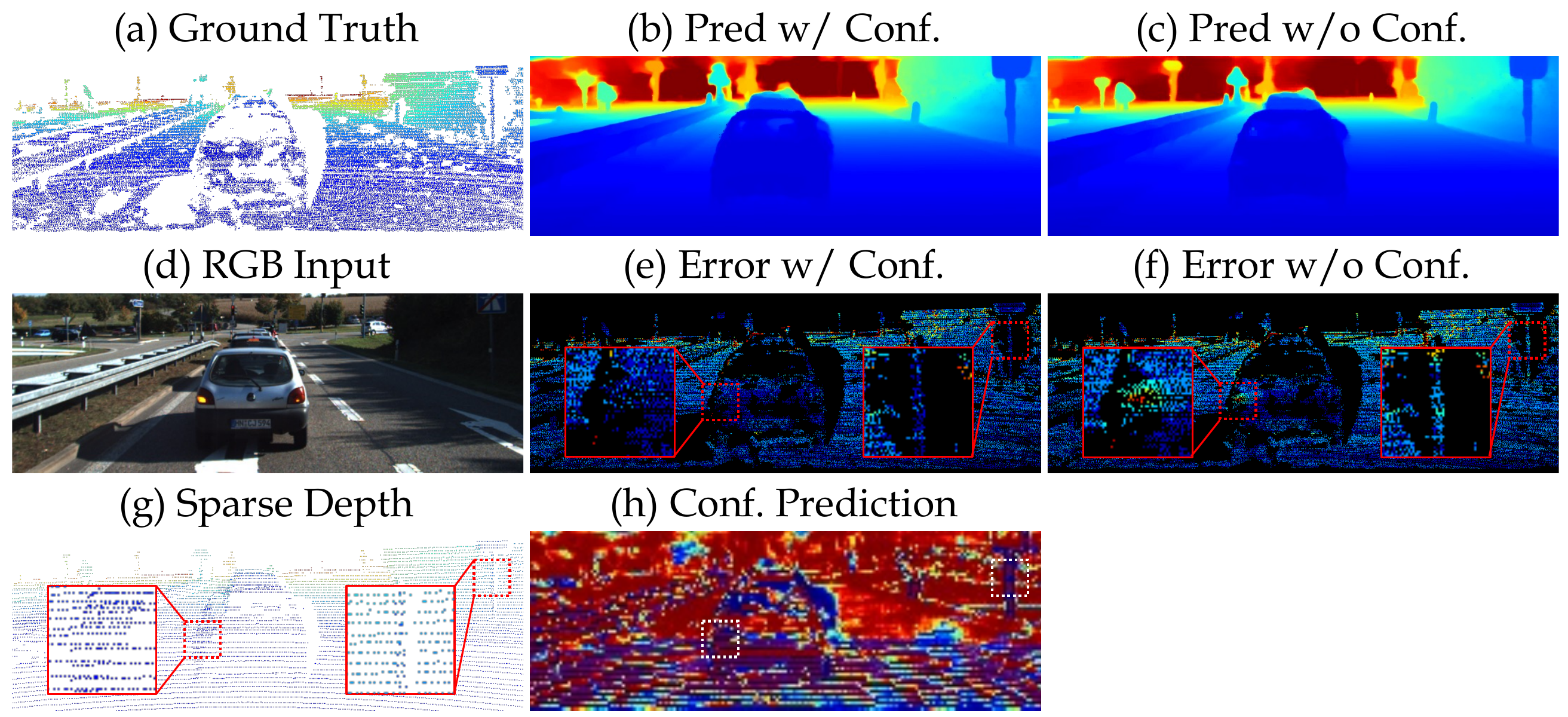}
  \caption{Our model learns a confidence map (h) to filter out the noisy sparse observations in (g). Compare (e) to (f), the model with confidence prediction is more accurate.}
  \label{fig:conf_input}
  \vspace{-5mm}
\end{figure}

The sparse depth observations in the KITTI~\cite{kittidc} dataset contain bleeding artifacts due to the baseline between the camera and the Lidar sensor (see \cref{fig:conf_input} (g) for an example). This has been observed in several previous works~\cite{conti2022unsupervised,qiu2019deeplidar,park2020non}. This issue is especially critical for OGNI-DC, since the noisy inputs directly affect the outputs of DDI and the error cannot be corrected. Therefore, we predict a confidence map $\mathbf{\hat{C}} \in [0,1]^{\frac{H}{4} \times \frac{W}{4}}$ for the sparse depth observations from the 1/4 resolution feature with a Conv-Sigmoid layer. $\mathbf{\hat{C}}$ works by down-weighting the observation energy term for the noisy pixels in the optimization problem, \ie,

\begin{equation}
\label{eq:least_square_matrix_form_conf}
\renewcommand\arraystretch{1.3}
\overline{\hat{\mathbf{D}}} = \argmin_{\overline{\mathbf{D}}} 
\left\| 
\underbrace{\begin{pmatrix}
\mathbf{\overline{\Delta_x}} \\
\mathbf{\overline{\Delta_y}} \\
\operatorname{diag}\left(
\sqrt{\alpha} \cdot \overline{\sqrt{{\mathbf{\hat{C}}}}} \cdot \overline{\mathbf{M}}
\right) \\
\end{pmatrix}}_{\mathbf{A}}
\mathbf{\overline{D}} - 
\underbrace{\begin{pmatrix}
\overline{\mathbf{\hat{G}^x}} \\
\overline{\mathbf{\hat{G}^y}} \\
\sqrt{\alpha} \cdot  \overline{\sqrt{{\mathbf{\hat{C}}}}} \cdot \overline{\mathbf{M}} \cdot \overline{\mathbf{O}} \\
\end{pmatrix}}_{\mathbf{b}}
\right\| _2^2.
\end{equation}

Applying $\mathbf{\hat{C}}$ in the forward pass is straightforward. However, since we don't have ground truth for $\mathbf{\hat{C}}$, we must learn $\mathbf{\hat{C}}$ directly from the loss on the integrated depth map. Therefore, we have to compute the gradient of $\overline{\hat{\mathbf{D}}}$ with respect to $\mathbf{\hat{C}}$. This can be done by applying the chain rule (remember $\overline{\hat{\mathbf{D}}} = (\mathbf{A}^\intercal \mathbf{A})^{-1}\mathbf{A}^\intercal \mathbf{b}$):

\begin{gather}
\label{eq:backward_confidence}
\renewcommand\arraystretch{1.4}
(\mathbf{A}^\intercal \mathbf{A}) \cdot \overline{\hat{\mathbf{D}}} = \mathbf{A}^\intercal \mathbf{b} 
\;\Rightarrow\; 
\begin{matrix}
    \partial (\mathbf{A}^\intercal \mathbf{A}) \\ \hline
    \partial \overline{\mathbf{\hat{C}}}
\end{matrix} 
\cdot \overline{\hat{\mathbf{D}}} 
+ (\mathbf{A}^\intercal \mathbf{A}) \cdot 
\begin{matrix}
    \partial \overline{\hat{\mathbf{D}}} \\ \hline
    \partial \overline{\mathbf{\hat{C}}} 
\end{matrix} 
=
\begin{matrix}
    \partial (\mathbf{A}^\intercal\mathbf{b}) \\ \hline
    \partial \overline{\mathbf{\hat{C}}}
\end{matrix}
\\
\renewcommand\arraystretch{1.4}
\Rightarrow \;
\begin{matrix}
    \partial \overline{\hat{\mathbf{D}}} \\ \hline
    \partial \overline{\mathbf{\hat{C}}} 
\end{matrix} 
=
(\mathbf{A}^\intercal \mathbf{A} )^{-1} \left(
\begin{matrix}
    \partial (\mathbf{A}^\intercal\mathbf{b}) \\ \hline
    \partial \overline{\mathbf{\hat{C}}}
\end{matrix}
\;-\;
\begin{matrix}
    \partial (\mathbf{A}^\intercal \mathbf{A}) \\ \hline
    \partial \overline{\mathbf{\hat{C}}}
\end{matrix} 
\cdot \overline{\hat{\mathbf{D}}} 
\right),
\\
\renewcommand\arraystretch{1.4}
\enspace
\text{where} \;
\enspace
\begin{matrix}
    \partial (\mathbf{A}^\intercal\mathbf{b}) \\ \hline
    \partial \overline{\mathbf{\hat{C}}}
\end{matrix}
\;
= \operatorname{diag}\left(\alpha \cdot \overline{\mathbf{M}} \cdot \overline{\mathbf{O}} \right)
, \enspace
\left[
\begin{matrix}
    \partial (\mathbf{A}^\intercal \mathbf{A}) \\ \hline
    \partial \overline{\mathbf{\hat{C}}}
\end{matrix} 
\right]_{ijk}
= \alpha \cdot \overline{\mathbf{M}}_i \cdot \mathbbm{1}_{i=j=k}.
\end{gather}

We demonstrate the effectiveness of the predicted confidence map in \cref{fig:conf_input}. \cref{fig:conf_input} (h) shows that confidence is high (red) in most areas while being low (blue) at the noisy regions on the car and the traffic sign. Comparing \cref{fig:conf_input} (e) with \cref{fig:conf_input} (f), the errors are greatly reduced in those areas with predicted confidence map. The results show that our model successfully learned to filter out the noisy inputs even without the ground truth.

\section{Network Architecture}

\begin{figure}[ht]
\vspace{-5mm}
  \centering
  \includegraphics[width=0.9\linewidth]{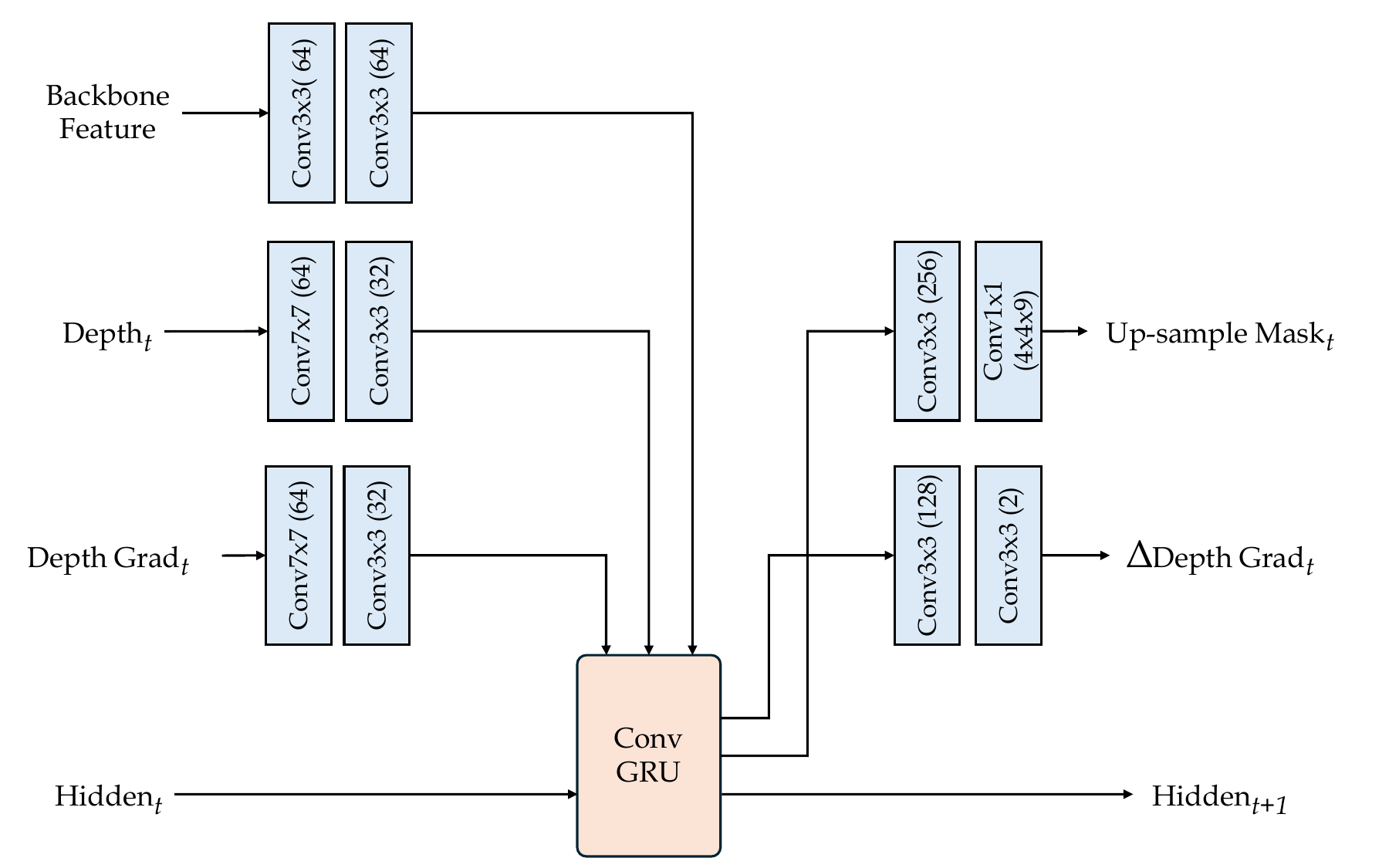}
  \caption{The detailed architecture of our update unit. }
  \label{fig:update unit}
\end{figure}

We use the CompletionFormer~\cite{zhang2023completionformer} as our backbone. CompletionFormer is a U-Net-like architecture with a series of down-sample and up-sample layers. The architecture of our update block is illustrated in \cref{fig:update unit}. We use the same ConvGRU as RAFT~\cite{teed2020raft}. Please refer to \cite{teed2020raft} for details.

\section{More Visualizations}

\subsection{NYUv2 Sparsity Generalization}

\begin{figure}[h]
\vspace{-5mm}
  \centering
  \includegraphics[width=0.8\linewidth]{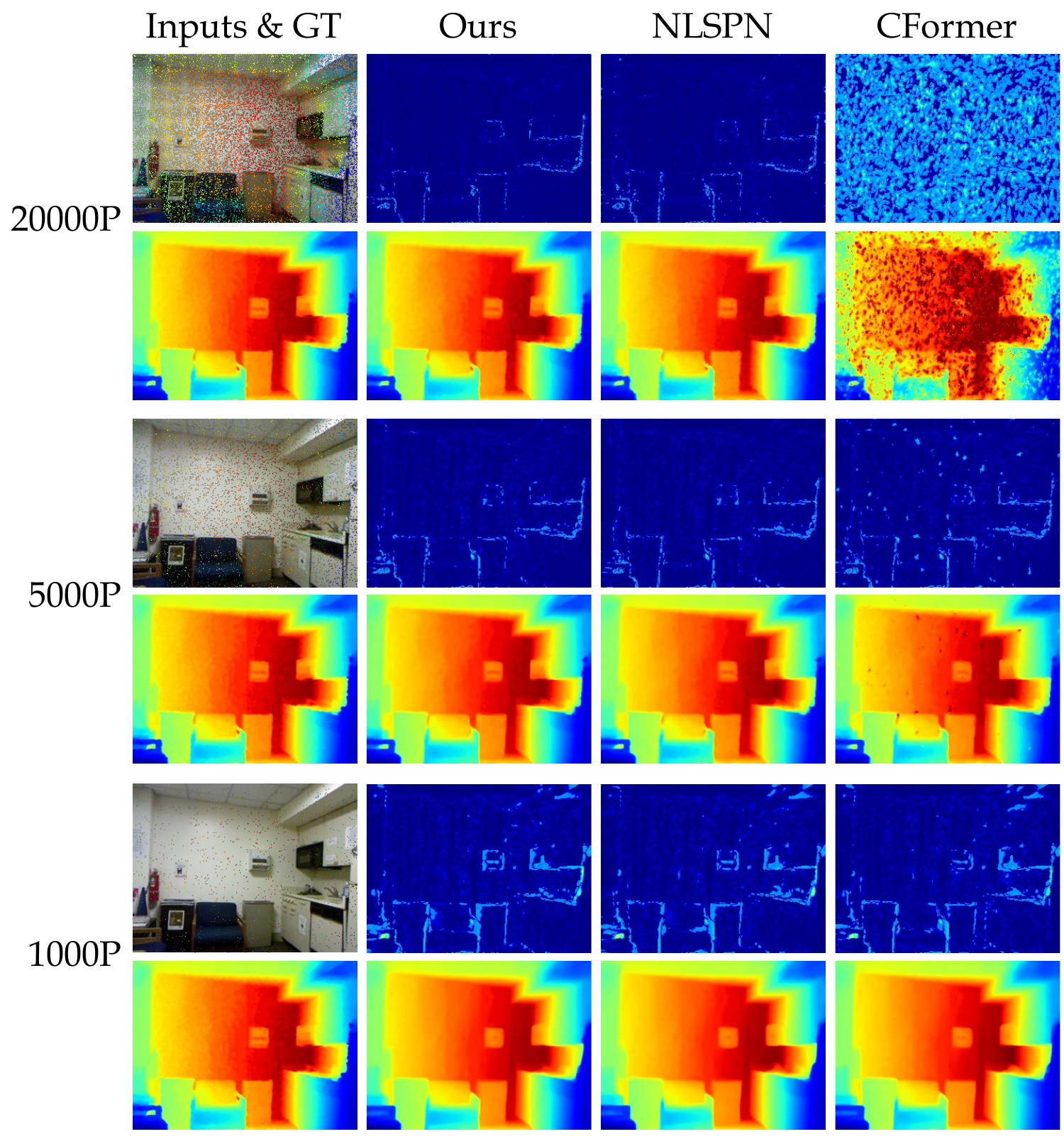}
  \caption{Generalization to denser inputs on NYUv2~\cite{nyuv2}. For each sparsity level, the first row is the inputs/error maps, and the second row is the gt/predicted depths. Our model works better than baselines although no models are trained on these sparsity levels.}
  \label{fig:nyu_500_20000}
\end{figure}

\begin{figure}[t]
  \centering
  \includegraphics[width=0.96\linewidth]{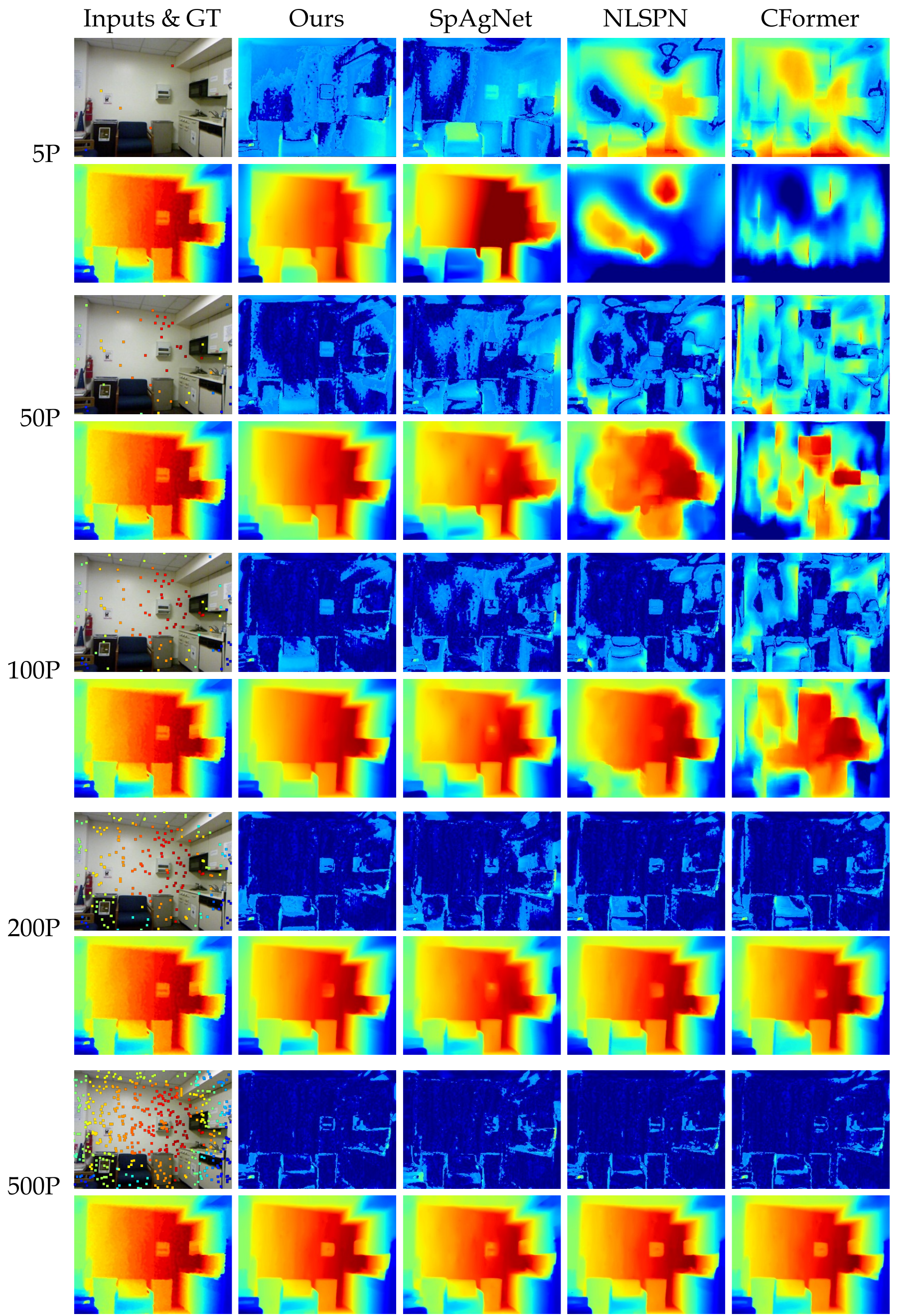}
  \caption{Generalization to sparser inputs on NYUv2~\cite{nyuv2}. For each sparsity level, the first row is the inputs/error maps, and the second row is the gt/predicted depths. Our model consistently outperforms baselines under all sparsity levels.}
  \label{fig:nyu_sparse_5_500}
\end{figure}

We qualitatively compare our method's generalization ability to different sparsity levels on the NYUv2~\cite{nyuv2} dataset with NLSPN~\cite{park2020non}, CFormer~\cite{zhang2023completionformer}, and SpAgNet~\cite{conti2023sparsity}.
Results are shown in \cref{fig:nyu_500_20000} and \cref{fig:nyu_sparse_5_500}. This image is the first one in the NYUv2~\cite{nyuv2} test set and is \textbf{not cherry-picked}. Our model works better than baselines under all sparsity levels. 

\subsection{KITTI Sparsity Generalization}

\begin{figure}[ht]
  \centering
  \includegraphics[width=\linewidth]{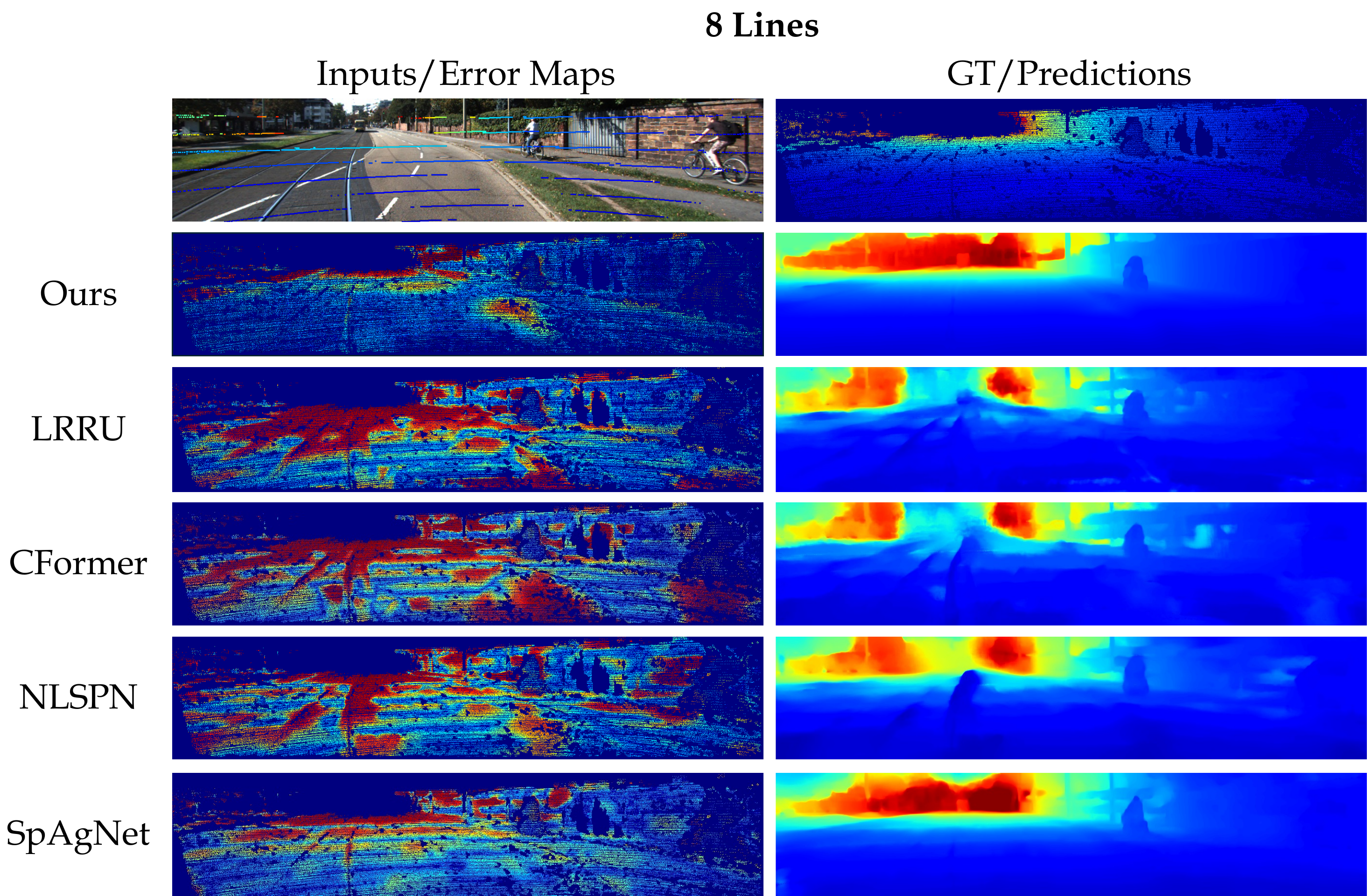}
  
  \vspace{1cm}
  
  \includegraphics[width=\linewidth]
  {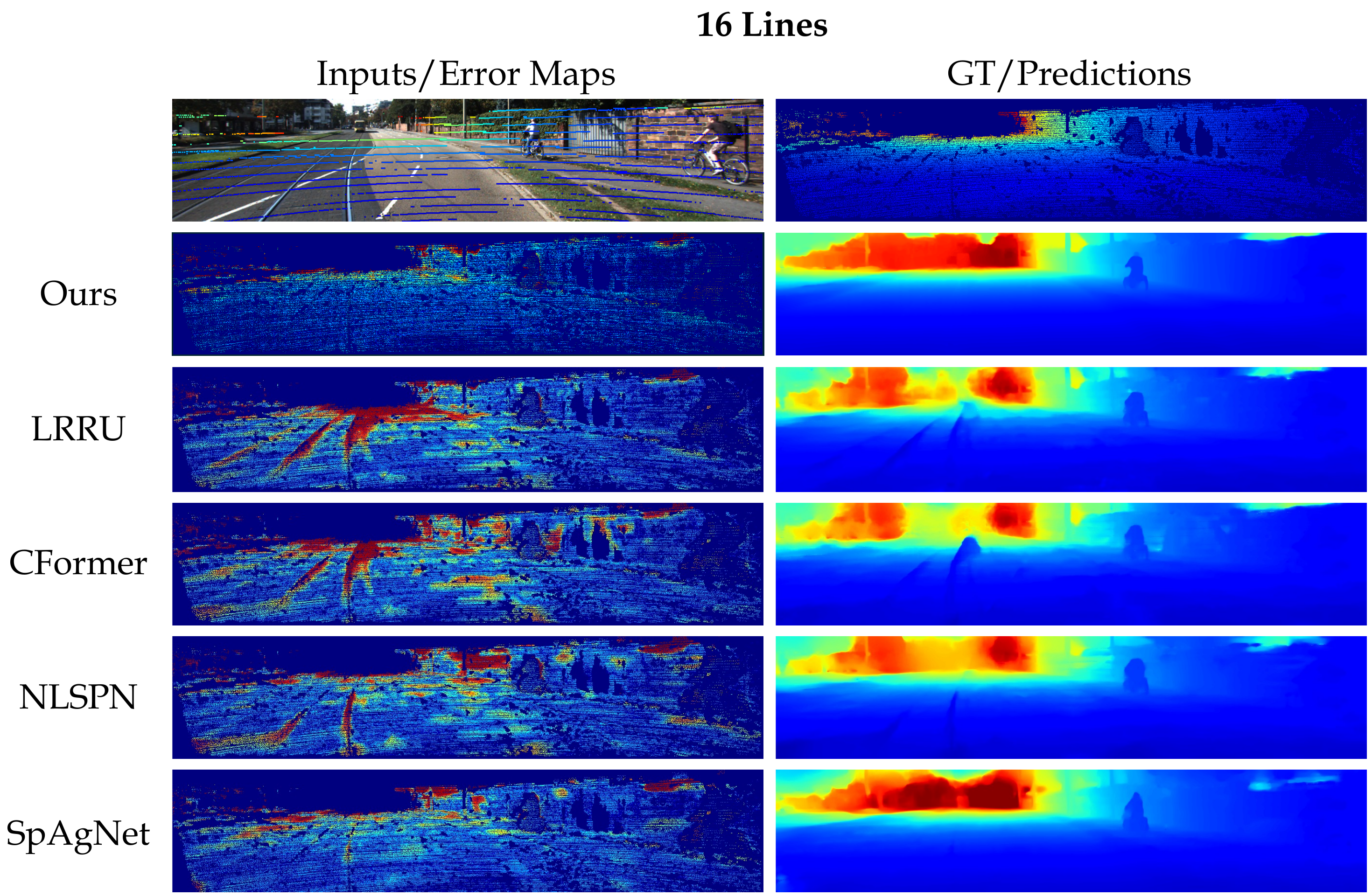}
  \caption{KITTI~\cite{kittidc} results with 8 and 16 lines inputs. Red colors mean larger errors.}
  \label{fig:kitti_8_16}
\end{figure}

\begin{figure}[ht]
  \centering
  \includegraphics[width=\linewidth]{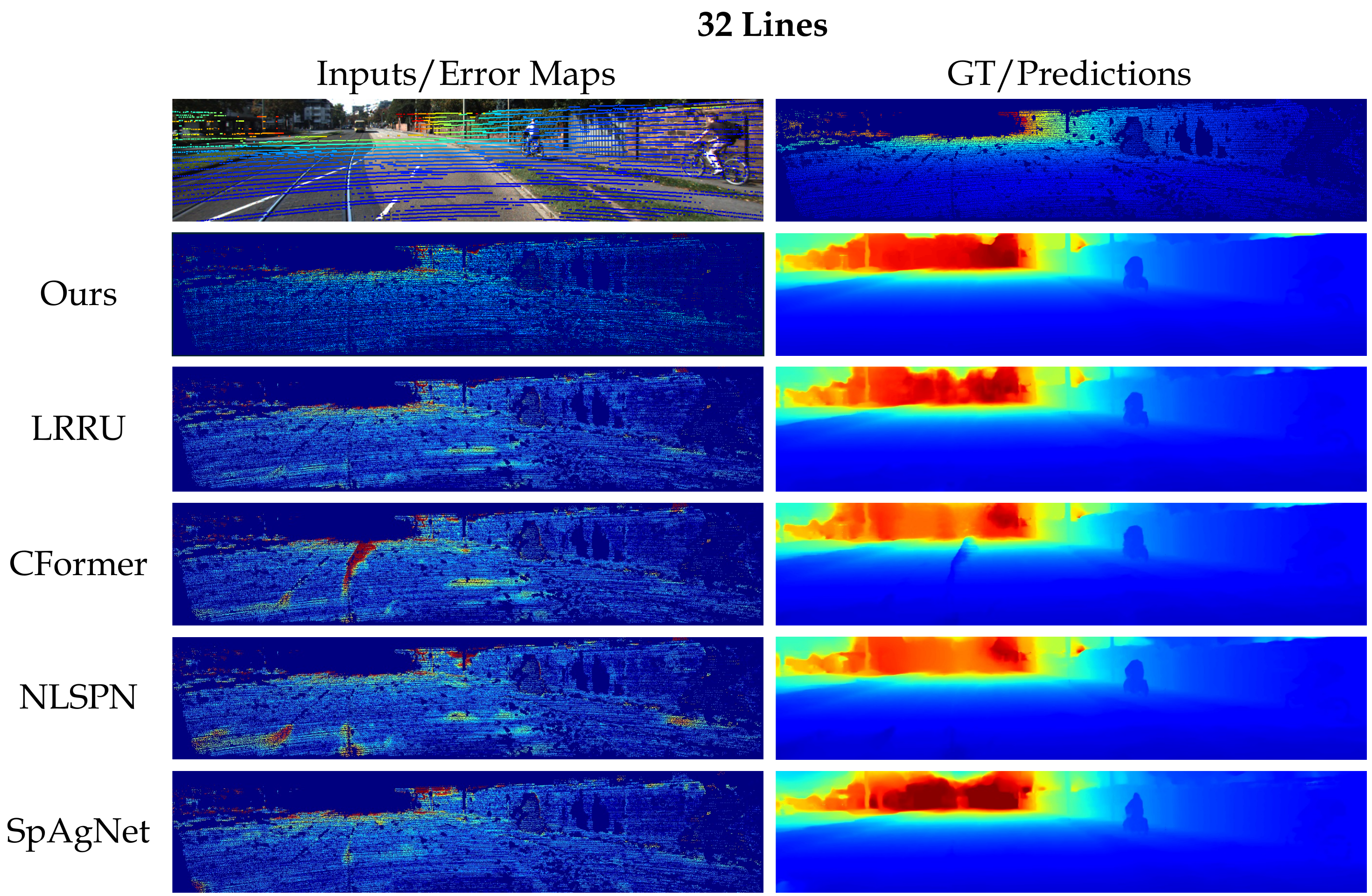}

  \vspace{1cm}
  
  \includegraphics[width=\linewidth]{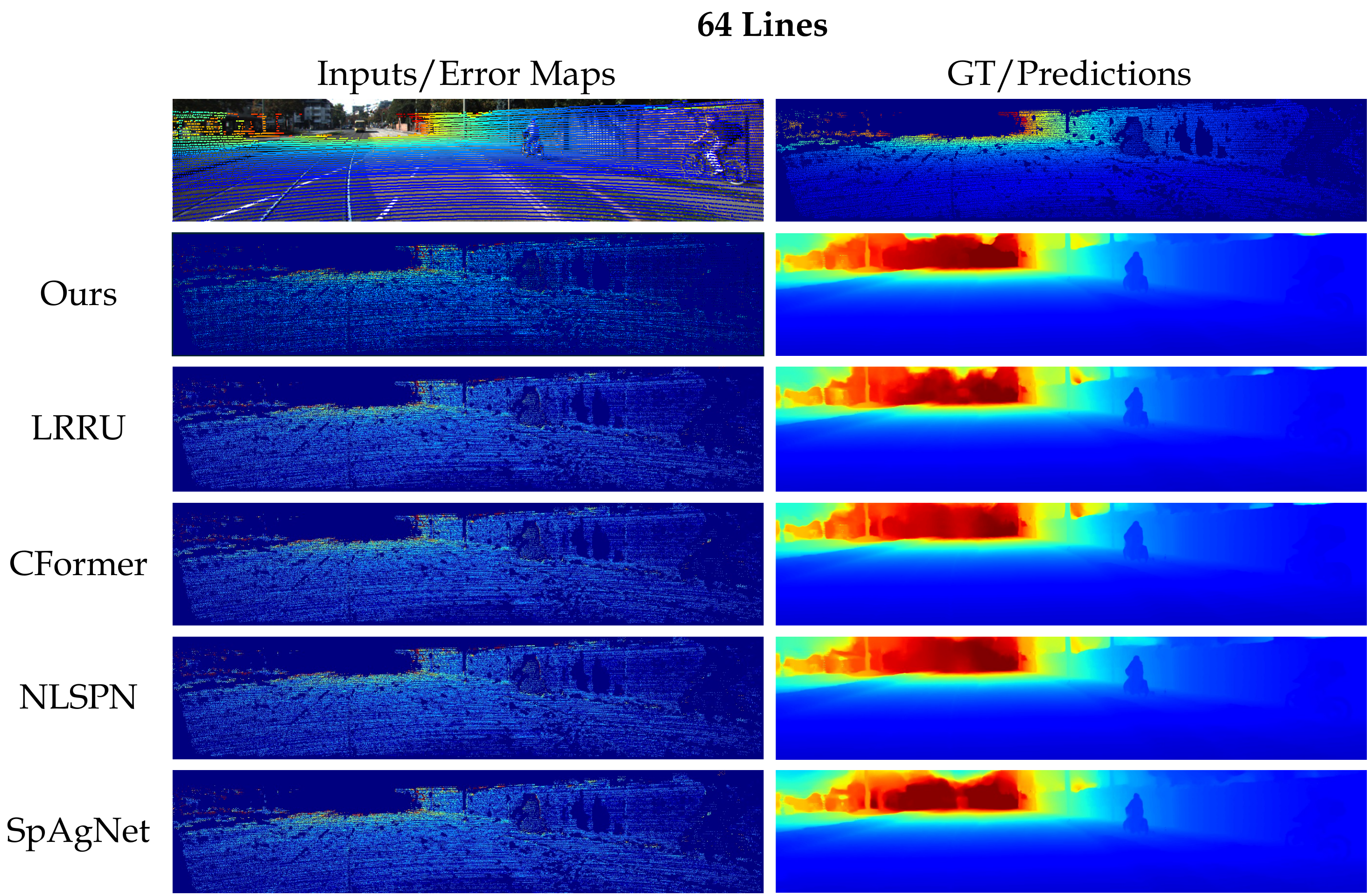}
  \caption{KITTI~\cite{kittidc} results with 32 and 64 lines inputs. Red colors mean larger errors.}
  \label{fig:kitti_32_64}
\end{figure}

Results are shown in \cref{fig:kitti_8_16} and \cref{fig:kitti_32_64}. This is the first image in the KITTI validation set and is \textbf{not cherry-picked}. While all methods do equally well with 64-Lines input, our method is significantly better when the inputs are sparser.

\clearpage
\section{Dataset Descriptions}

\littlesection{NYUv2.} NYUv2~\cite{nyuv2} contains \numprint{45205} training images, \numprint{2379} validation images, and \numprint{654} test images from 464 indoor scenes. Dense depth maps are collected with the Microsoft Kinect sensor, and 500 points are randomly sampled to provide sparse observations. Following previous works~\cite{park2020non,zhang2023completionformer}, we resize the original $480\times640$ images to $240\times320$ and then center-crop to $228\times304$.

\littlesection{KITTI.} The KITTI depth completion dataset~\cite{kittidc} contains \numprint{86898} training images, \numprint{1000} selected validation images, and \numprint{1000} online test images. Images and depths are collected from an autonomous driving vehicle with a Velodyne HDL-64E Lidar sensor. All images have resolution $352\times1216$. Following previous works~\cite{park2020non,zhang2023completionformer}, during training and validation, the images are bottom-cropped to $240\times1216$ as no Lidar points are available in the sky areas.

\littlesection{VOID.} The VOID~\cite{void} dataset contains 56 sequences of indoor scenes. Depth ground truths are collected with an Intel RealSense D435i camera, and sparse observations are from a visual odometry system at 3 different sparsity levels, \ie, \numprint{1500}, 500, and 150 points, corresponding to 0.5\%, 0.15\%, and 0.05\% density. Each test split contains 800 images at $480\times640$ resolution.

\littlesection{DDAD.} DDAD~\cite{ddad} is an autonomous driving dataset with depth ground truth captured by a long-range, high-resolution Luminar-H2 Lidar. Following the split and pre-processing of VPP4DC~\cite{bartolomei2023revisiting}, we evaluate on \numprint{3950} images under $1216\times1936$ resolution captured by the front-viewing camera. We randomly sample about 20\% points as the input sparse depth, resulting in $\sim0.21\%$ density.

\end{document}